\documentclass[journal]{IEEEtran}
\usepackage{tikz,xcolor}

\usepackage{amsmath,amsfonts}
\usepackage{algorithmic}
\usepackage{algorithm}
\usepackage{array}
\usepackage[caption=false,font=normalsize,labelfont=sf,textfont=sf]{subfig}
\usepackage{textcomp}
\usepackage{stfloats}
\usepackage{url}
\usepackage{verbatim}
\usepackage{graphicx}
\usepackage{cite}
\usepackage{multirow}
\usepackage{bm}
\usepackage{balance}
\usepackage{hyperref}
\hypersetup{
    colorlinks=true,     
    linkcolor=blue,      
    citecolor=blue,      
}

\definecolor{lime}{HTML}{A6CE39}
\DeclareRobustCommand{\orcidicon}{
\begin{tikzpicture}
\draw[lime, fill=lime] (0,0)
circle[radius=0.16]
node[white]{{\fontfamily{qag}\selectfont \tiny \.{I}D}};
\end{tikzpicture}
\hspace{-2mm}
}
\foreach \x in {A, ..., Z}{%
\expandafter\xdef\csname orcid\x\endcsname{\noexpand\href{https://orcid.org/\csname orcidauthor\x\endcsname}{\noexpand\orcidicon}}
}

\begin{document}

\title{Direction-Oriented Visual-semantic Embedding Model for Remote Sensing Image-text Retrieval}

\author{
Qing Ma\hspace{-1.5mm}\orcidC{},~\IEEEmembership{Member,~IEEE,},
Jiancheng Pan\hspace{-1.5mm}\orcidA{},~\IEEEmembership{Student Member,~IEEE},
Cong Bai\hspace{-1.5mm}\orcidB{},~\IEEEmembership{Member,~IEEE,}

\thanks{Manuscript received 12 October 2023; revised 25 January 2024 and 28 March 2024; accepted 20 April 2024. Date of publication 23 April 2024; date of current version 8 May 2024. This work is partially supported by Natural Science Foundation of China under Grant No. 61976192, and Zhejiang Provincial Natural Science Foundation of China under Grant No. LR21F020002 and National Key Research and Development Program of China (No. 2018YFE0126100). (Corresponding authors: Cong Bai.)}
\thanks{Qing Ma and Jiancheng Pan contributed equally. Qing Ma is with the College of Science, Zhejiang University of Technology, Hangzhou 310023, China. Jiancheng Pan and Cong Bai are with the College of Computer Science and Technology, Zhejiang University of Technology, Hangzhou 310023, China (e-mail: congbai@zjut.edu.cn).}}

\IEEEpubid{0000--0000/00\$00.00~\copyright~2021 IEEE}

\maketitle

\begin{abstract}
Image-text retrieval has developed rapidly in recent years. However, it is still a challenge in remote sensing due to visual-semantic imbalance, which leads to incorrect matching of non-semantic visual and textual features. To solve this problem, we propose a novel Direction-Oriented Visual-semantic Embedding Model (DOVE) to mine the relationship between vision and language. \textcolor{black}{Our highlight is to conduct visual and textual representations in latent space, directing them as close as possible to a redundancy-free regional visual representation.} Concretely, a Regional-Oriented Attention Module (ROAM) adaptively adjusts the distance between the final visual and textual embeddings in the latent semantic space, oriented by regional visual features. Meanwhile, a lightweight Digging Text Genome Assistant (DTGA) is designed to expand the range of tractable textual representation and enhance global word-level semantic connections using less attention operations. Ultimately, we exploit a global visual-semantic constraint to reduce single visual dependency and serve as an external constraint for the final visual and textual representations. The effectiveness and superiority of our method are veriﬁed by extensive experiments including parameter evaluation, quantitative comparison, ablation studies and visual analysis, on two benchmark datasets, RSICD and RSITMD.
\end{abstract}


\section{Introduction}
\begin{figure}[htbp]
  \centering
  \includegraphics[width=0.95\linewidth]{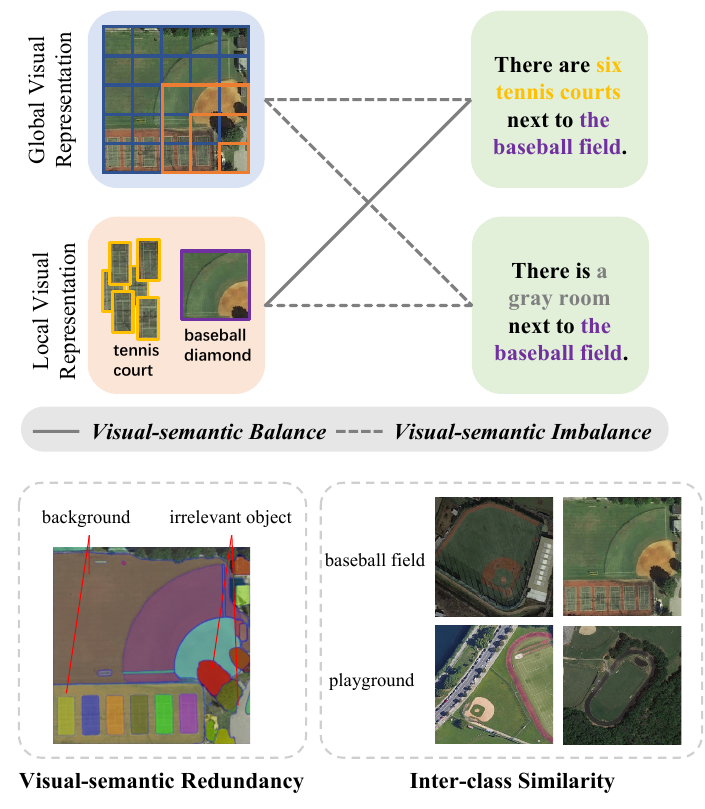}
  \caption{(a): Visual-semantic balance and visual-semantic imbalance. (b): Two main factors cause visual-semantic imbalance, visual-semantic redundancy and inter-class similarity.}
  \label{fig:fig1}
\end{figure}

\IEEEPARstart{I}{mage-text} retrieval \cite{qu2021dynamic,zhang2022image} has received much attention from researchers as a typical task in multimodal learning, querying a text (or image) with an image (or text), to mine the deep association between vision and language. Similarly, image-text retrieval in remote sensing \cite{yuan2022remote,mi2022knowledge} has become an important research topic, playing an important role in resource exploration, disaster monitoring \cite{chi2016big, joyce2009review}, and remote sensing vision-language (RSVL) tasks such as remote sensing image captioning (RSIC) \cite{cheng2022nwpu, zhao2021high, liu2022remote}. Compared with traditional image-text retrieval based on natural images, remote sensing image-text retrieval has a visual-semantic imbalance problem that leads to the incorrect matching of non-semantic visual features and textual features, as shown in Fig. \ref{fig:fig1}(a). Fig. \ref{fig:fig1}(b) shows two main factors causing this problem: 1) \textbf{Visual-semantic redundancy}. Remote sensing images contain many small-scale semantic objects, whose semantic representation is vulnerable to interference from non-semantic components (eg. background and irrelevant objects); 2) \textbf{Inter-class similarity}. The apparent similarity of images with different scenes can easily lead to inaccurate visual semantic representation.

A better visual representation is a key to successful remote sensing image-text retrieval\textcolor{black}{\cite{li2022vision,10197537,9706143}.} According to the visual representation methods, current remote sensing image-text retrieval methods are either global visual representation-based or global and local visual representation-based.\IEEEpubidadjcol The global visual features are usually extracted by mapping the final layers of output from CNNs \cite{lecun1998gradient} or Vision Transformer \cite{dosovitskiy2020image} into the final visual embeddings. \textbf{Global visual representation-based method} \cite{mao2018deep, abdullah2020textrs, lv2021fusion, cheng2021deep, mi2022knowledge} uses only global visual features as final visual representation. Some approaches \cite{mao2018deep, abdullah2020textrs} directly encode data of different modalities into corresponding features and measure the similarity in the latent space, which is feasible for remote sensing image-text matching, but they do not pay more attention to semantic redundancy of these data. To solve the problem of coarse-grained textual descriptions, Might et al. \cite{mi2022knowledge} proposed a knowledge-aware method to get relevant information from an external knowledge graph. Although global visual features contain most of the semantics of the image, it often contains a large amount of redundancy. For example, the overlap of multiple perceptual fields in the deep features of CNNs causes them to contain many irrelevant semantic features. Local visual features are generally extracted using Faster R-CNN \cite{ren2015faster}, or YOLO \cite{redmon2016you} for object detection. Since current remote sensing image-based object detection algorithms do not work well when processing low-resolution and small-scale object images, using only local visual features will cause a severe visual-semantic imbalance. \textbf{Global and local visual representation-based method} \cite{yuan2022remote, zhang2023hypersphere} generally fuses global and local visual features as a final visual representation using the attention mechanism. Yuan et al. \cite{yuan2022remote} used a graph convolutional network (GCN) \cite{kipf2016semi} to enhance the relationship of salient objects, and designed an attention-based module to dynamically fuse multilevel visual information. Zhang et al. \cite{zhang2023hypersphere} proposed a hypersphere-based visual semantic alignment (HVSA) network via curriculum learning to solve the characteristics of data distribution and the varying difficulty levels of different sample pairs. These method rely on a good fusion strategy to enhance visual representation, and ignores the deeper relationship between regional visual features and textual features. The above two types of methods have achieved promising success in remote sensing image-text retrieval, but have paid less attention to the discrepancy between vision and semantics. For visual-semantic imbalanced image-text pairs, over-reliance on local visual features may lead to alignments dominated by a single object semantic or non-semantic component, while over-reliance on global visual features makes the visual representation vulnerable to extensive redundancy. 

\begin{figure*}[ht]
  \centering
  \includegraphics[width=0.95\linewidth]{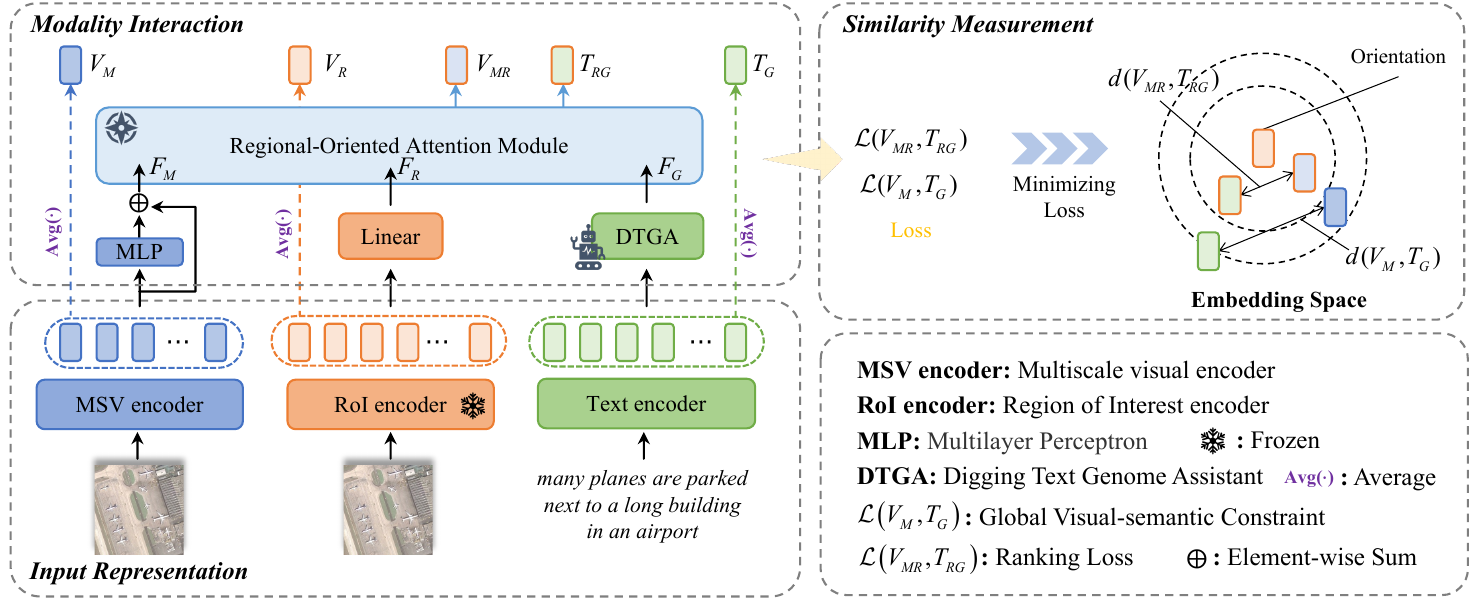}
  \caption{Schematic illustration of DOVE model. Internal constraint with the regional visual embedding as orientation and external boundary of global visual-semantic constraint allows matching visual and textual embeddings to approximate each other in the latent embedding space.}
  \label{fig:fig2}
\end{figure*}

To address visual-semantic imbalance problem, we propose a \textbf{D}irection-\textbf{O}riented \textbf{V}isual-semantic \textbf{E}mbedding Model (DOVE) to achieve fine-grained alignment of remote sensing images and text. \textcolor{black}{Unlike general image-text retrieval methods, the DOVE represents the learning of de-biased representations by taking the regional visual features as references to direct the final visual and textual representations to be as close as possible to the relatively redundancy-free regional visual representations.} The DOVE consists of the input representation, modality interaction, and similarity measurement, as shown in Fig. \ref{fig:fig2}. In the input representation part, a Digging Text Genome Assistant (DTGA) interacts with the forward and backward hidden outputs of the Gated Recurrent Unit (GRU) \cite{chung2014empirical} to obtain enhanced textual features. In the modality interaction part, a Regional-Oriented Attention Module (ROAM) explores the deep connection between vision and language, oriented by regional visual features. A global visual-semantic constraint is employed as an external constraint for the final visual and textual representations in similarity measurement and reduce single visual dependency. Experiments on the  RSICD \cite{lu2017exploring} and RSITMD \cite{yuan2022exploring} datasets showed that our method has a significant advantage over current state-of-the-art methods, with great improvement on most metrics.

The main contributions of our work are as follows:
\begin{itemize}
\item We propose a novel remote sensing image-text retrieval model DOVE, which can solve the problem of visual-semantic imbalance and strengthen the association between vision and language to achieve fine-grained alignment of images and text;
\item The DTGA module, based on a dual-branch symmetrical structure, is proposed to enhance textual representation with global word-level contextual relationships and effectively mitigate visual-semantic imbalance by improving textual semantic representation;
\item To explore the internal connection between vision and language, the ROAM module is designed to adaptively adjust the distance between the final visual and textual embeddings in the latent embedding space, using regional visual features as orientation.
\item A global visual-semantic constraint acts as an external constraint for the final visual and textual representations and alleviates the single visual dependency.
\end{itemize}

\section{RELATED WORK}
\subsection{Remote Sensing Image-Text Retrieval}
In recent years, image-text retrieval based on remote sensing has also gradually received the attention of some researchers. The existing remote sensing image-text retrieval can be roughly divided into two categories for the modal interaction \cite{qu2021dynamic}: 1) intra-modal interaction method and 2) inter-modal interaction method.

Intra-modal interaction method \cite{abdullah2020textrs, mi2022knowledge, yuan2022remote, zhang2022transformer, zhang2023hypersphere} performs information interaction between homogeneous modalities before entering the latent semantic space. Abdullah et al. \cite{abdullah2020textrs} fused five corresponding images and text by an averaging fusion strategy to achieve remote sensing image-text retrieval. Might et al. \cite{mi2022knowledge} extended the textual semantic scope using knowledge graphs to obtain a more robust textual representation. Yuan et al. \cite{yuan2022remote} proposed a Multi-Level Information Dynamic Fusion module that dynamically fuses global and local visual information to obtain a salient visual representation. Zhang et al. \cite{zhang2022transformer} proposed a  module to reconstruct decoupled features, ensuring that the amount of information in the features was maximally preserved, and employed orthogonality constraints and adversarial learning to optimize model. Zhang et al. \cite{zhang2023hypersphere} introduced a Hypersphere-Based Visual Semantic Alignment (HVSA) network, leveraging curriculum learning to address the challenges arising from variations in data distribution and the diverse difficulty levels among different sample pairs. Although these methods can obtain an independent modal representation, they focus only on the modality itself, and not on the subtle connections between modalities.

Inter-modal interaction method \cite{lv2021fusion, cheng2021deep, yuan2022exploring, yuan2021lightweight} performs information interaction between different modalities before entering the latent semantic space. Lv et al. \cite{lv2021fusion} designed a cross-modal fusion network to capture the fused information between modalities and transfer it to supervised modal representation through knowledge distillation. Cheng et al. \cite{cheng2021deep} used an attention mechanism to enhance relationship between images and text, and designed a gating function to obtain discriminative visual and textual features. Yuan et al. \cite{yuan2022exploring} used a multiscale visual self-attention module to extract salient features of images, and used visual features to guide textual representation. Yuan et al. \cite{yuan2021lightweight} proposed a supervised optimization method based on knowledge distillation to maintain a lightweight retrieval models. These methods can mine the association relationships between different modalities and obtain the most valuable semantic features.

Although many researchers have considered using intra- and inter-modal interactions to improve retrieval performance, they have ignored the visual-semantic imbalance caused by remote sensing image characteristics. A significant challenge of remote sensing image-text retrieval is to design a network structure that can fully use the correlations between different modalities and effectively solve the visual-semantic imbalance.
\subsection{Attention Mechanism}
The attention mechanism, a breakthrough technology in artificial intelligence, is widely used in cross-modal image-text retrieval, which reduces the computational burden of high-dimensional input and focuses more on the representation of salient information. A Dual Attention Network was proposed by Nam et al. \cite{nam2017dual} to concentrate on particular areas in images and words. To find the whole latent alignments and infer image-text similarity, Lee et al. \cite{lee2018stacked} presented Stacked Cross Attention using both image regions and sentence words as context to achieve fine-grained alignment. Wang et al. \cite{wang2019camp} presented a Cross-modal Adaptive Message Passing Model to adaptively explore interactions between images and sentences for image-text matching. Following this work, many approaches \cite{ji2019saliency, zhang2020context, wei2020multi, ji2021step} have been used to mine the potential connections between images and text through cross-modal interaction between vision and language. Li et al. \cite{li2022action} used transformer-based cross-modal attention module to achieve image-text retrieval, which incorporates action-similar sentences from the memory bank to improve action-aware embedding.

For various modality attention structures, we propose a universal modality attention module, ROAM (see Section \ref{sectionAttention}), which adaptively adjusts the visual and linguistic representation in the semantic space in accordance with regional visual features, and can fully use the information exchange between modalities to improve modal semantic representation.

\section{METHODOLOGY}
Fig. \ref{fig:fig2} shows the proposed DOVE model. We focus on four aspects: 1) input representation for visual and textual modalities; 2) Digging Text Genome Assistant (DTGA) to enhance text fine-grained representation; 3) Regional-Oriented Attention Module (ROAM) to mine the deep connection between vision and language; and 4) objective function for the alignment of images and text.
\subsection{Input Representation}
\subsubsection{Visual Representation}
Previous studies \cite{yuan2022remote,ji2019saliency} have demonstrated that to use only global visual features is not a good method to achieve image-text retrieval. Unlike the general natural image-based image-text retrieval approach, we utilize a multiscale visual (MSV) encoder to extract multiscale visual features, using a pre-trained ResNet-50 \cite{he2016deep} on the AID dataset \cite{xia2017aid} as the backbone. We detect salient regions by the Region of Interest (RoI) encoder \cite{ding2019learning}, using ResNet-50 as the backbone.

Given an image input \bm{$I$}, the multiscale visual features $\bm{M}_v=[\bm{v}_1, \bm{v}_2,...,\bm{v}_{N_m}]^{\mathrm{T}} \in \mathbb{R}^{N_m \times d}$ are obtained by MSV encoder, and region features $\bm{R}_v=[\bm{u}_1, \bm{u}_2,...,\bm{u}_{N_r}]^{\mathrm{T}} \in \mathbb{R}^{N_r \times d/2}$ are acquired by RoI encoder. For an image with 256 $\times$ 256 resolution, we have $N_m=4$ (the semantic-level features from $layer1$, $layer2$, $layer3$, $layer4$ of the ResNet-50) and $N_r=36$. The final multiscale visual feature is obtained by the multilayer perceptron (MLP) module as
\begin{equation}
\begin{aligned}
\bm{F}_M = M L P\left(\bm{M}_v\right) + \bm{M}_v,
\end{aligned}
\end{equation}
where $\bm{F}_M \in \mathbb{R}^{N_m \times d}$ represents the multiscale visual features, and the region features are transformed by a fully-connected layer as
\begin{equation}
\begin{aligned}
\bm{F}_R = \bm{R}_v \bm{W}_r + \bm{b}_r,
\end{aligned}
\end{equation}
where $\bm{F}_R \in \mathbb{R}^{N_r \times d}$ represents the regional visual features, and $\bm{W}_r$ and $\bm{b}_r$ denote the weights and bias of the fully-connected layer, respectively.
\subsubsection{Textual Representation}
To explore the connection between vision and language, we perform feature extraction on the text. Given a text input \bm{$T$}, we first encode them into one-hot encoding $\{\bm{w}_1, \bm{w}_2,...,\bm{w}_{N_c}\}$, where $\bm{w}_i \in \mathbb{R}^{d}(i \in[1,N_c])$, and embed them into 300-dimensional vectors as $\bm{e}_i  = \bm{W}_e\bm{w}_i(i \in[1,N_c])$, where $\bm{W}_e$ is the parametric matrix of Glove \cite{pennington2014glove}. We feed these vectors into the bidirectional GRU \cite{chung2014empirical} to learn the contextual relationships between words,
\begin{equation}
\begin{aligned}
\bm{h}_i^f =G R U^f\left(\bm{e}_i, \bm{h}_{i-1}^f\right), \ \bm{\mathcal{H}}^f = [\bm{h}_1^f, \bm{h}_2^f,..., \bm{h}_{N_c}^f]^{\mathrm{T}},
\end{aligned}
\end{equation}
\begin{equation}
\begin{aligned}
\bm{h}_i^b =G R U^b\left(\bm{e}_i, \bm{h}_{i+1}^b\right), \ \bm{\mathcal{H}}^b = [\bm{h}_1^b, \bm{h}_2^b,..., \bm{h}_{N_c}^b]^{\mathrm{T}},
\end{aligned}
\end{equation}
where $\bm{h}_i^f$ and $\bm{h}_i^b$ respectively represent the hidden state of the forward and backward GRU from the $i$-th layer, with respective hidden layer outputs $\bm{\mathcal{H}}^f$ and $\bm{\mathcal{H}}^b$. Unlike most methods \cite{ji2019saliency, zhang2020context, yuan2022remote, pan2023reducing}, which directly average the outputs of the forward and backward hidden layers to obtain textual embedding, we use a strategy (DTGA) based on dual-flow gating to dynamically fuse them (see Section \ref{sectionDTGA}). We can get the word-level textual features $\bm{F}_G \in \mathbb{R}^{N_c \times d}$ as
\begin{equation}
\begin{aligned}
\bm{F}_G =D T G A\left(\bm{\mathcal{H}}^f, \bm{\mathcal{H}}^b\right).
\end{aligned}
\end{equation}
\begin{figure}[t]
  \centering
  \includegraphics[width=\linewidth]{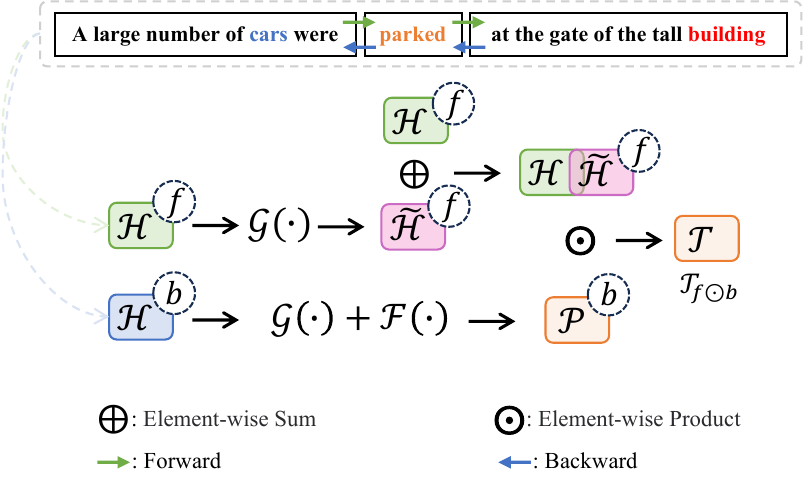}
  \caption{The use of text backward hidden layer features to mine text forward hidden layer features in DTGA module.}
  \label{fig:fig3}
\end{figure}
\subsection{Contextual Enhancement Strategy}\label{sectionDTGA}
\subsubsection{Gated Self-Attention (GA)}
It is challenging to use  only RNN \cite{jordan1997serial} or LSTM \cite{hochreiter1997long} to capture features with long-distance dependencies. The relationship between arbitrary words can be linked by using a gated self-attention \cite{zhao2018paragraph, qu2020context}. Let $\bm{\mathcal{X}} \in \mathbb{R}^{N_c \times d}$ represent the input of the GA module, through which we can obtain the attention feature $\tilde{\bm{\mathcal{X}}} \in \mathbb{R}^{N_c \times d}$ as
\begin{equation}\label{eqnGA}
\tilde{\bm{\mathcal{X}}} = \bm{\mathcal{G}}(\bm{\mathcal{X}}),
\end{equation}
where $\bm{\mathcal{G}}(\cdot)$ is the Gated Self-Attention. Concretely, we first get $\bm{\mathcal{Q}} \in \mathbb{R}^{N_c \times d}$, $\bm{\mathcal{K}} \in \mathbb{R}^{N_c \times d}$, and $\bm{\mathcal{V}} \in \mathbb{R}^{N_c \times d}$, which respectively denote the query, key, and value, and can be obtained as
\begin{equation}
\left\{\begin{array}{l}
\begin{aligned}
\bm{\mathcal{Q}} &= \bm{\mathcal{X}} \bm{W}_\mathcal{Q} + \bm{b}_\mathcal{Q}, \\
\bm{\mathcal{K}} &= \bm{\mathcal{X}} \bm{W}_\mathcal{K}+ \bm{b}_\mathcal{K}, \\
\bm{\mathcal{V}} &= \bm{\mathcal{X}} \bm{W}_\mathcal{V}+ \bm{b}_\mathcal{V},
\end{aligned}
\end{array}\right.
\end{equation}
where $\bm{W}_\mathcal{Q} (\bm{W}_\mathcal{K},\bm{W}_\mathcal{V})$ and $\bm{b}_\mathcal{Q} (\bm{b}_\mathcal{K},\bm{b}_\mathcal{V})$, respectively, are the the learnable weights and bias. Unlike general self-attention, we introduce a Gate Mechanism \cite{qu2020context} to deliver messages adaptively and suppress noisy or meaningless information. 
Let $\bm{G} \in \mathbb{R}^{N_c \times d}$ represent the gating activation value, which is calculated as
\begin{equation}
\bm{G}= \sigma\left( \left(\bm{\mathcal{Q}} \odot \bm{\mathcal{K}}\right) \bm{W}_A + \bm{b}_A \right),
\end{equation}
where $\bm{W}_A $ and $\bm{b}_A$ represent the learnable weights and bias, respectively, of a fully-connected layer; $\odot$ denotes the element-wise product operation; and $\sigma \left(\cdot \right)$ is the sigmoid function. Then we can get the activated $\bm{\mathcal{Q}}^{\prime} \in \mathbb{R}^{N_c \times d}$ and $\bm{\mathcal{K}}^{\prime} \in \mathbb{R}^{N_c \times d}$ as
\begin{equation}
\bm{\mathcal{Q}}^{\prime}=\bm{G} \odot \bm{\mathcal{Q}},
\end{equation}
\begin{equation}
\bm{\mathcal{K}}^{\prime}= \bm{G} \odot \bm{\mathcal{K}}.
\end{equation}
Finally, the scaled dot-product attention is
\begin{equation}
\tilde{\bm{\mathcal{X}}} =\operatorname{Softmax}\left(\frac{\bm{\mathcal{Q}}^{\prime} \bm{\mathcal{K}}^{\prime \mathrm{T}}}{\sqrt{d}}\right) \bm{\mathcal{V}},
\end{equation}
where $\operatorname{Softmax}\left(\cdot \right)$ represents the softmax function, which is performed for each row.
\subsubsection{Digging Text Genome Assistant (DTGA)}
GRU \cite{chung2014empirical} processes text by accepting a part of the input text (e.g., a word) at each time step and outputting a hidden state. The hidden state of GRU can be regarded as the model's understanding and memory of the input sequence up to the current time step. However, this memory is easily and gradually forgotten as the time step increments. And the forward and backward contextual relations focus on different long-distance information, their word-level contextual semantic links can be mined to improve retrieval performance. Many previous works \cite{ji2019saliency, zhang2020context, yuan2022remote, zhang2023hypersphere} directly compute their average valuess on the forward and backward hidden layer outputs of LSTM. However, simply averaging the forward semantics and backward semantics in image-text retrieval can damage the text semantic content.

Different words in a sentence have different probabilities of inference in the forward and backward directions. For the sentences ``many planes are parked next to a long building in an airport.'' and ``A large number of cars were parked at the gate of the tall building.'', the word ``\emph{parked}'' may be followed by either ``\emph{planes}'' or ``\emph{cars}'', while the probability of being followed by ``\emph{building} '' is quite high because ``\emph{building} '' and ``\emph{parked}'' are more strongly related in the text set than ``\emph{cars}'' and ``\emph{planes}''. And this relationship tends to have a stronger correlation representation in a single direction (forward or backward). Therefore it is necessary to mine the key features from the other direction features. The proposed lightweight DTGA module uses the forward and backward hidden layer outputs of the bidirectional GRU to obtain deeper relationships among words without much global attention operations. \textcolor{black}{In the DTGA module, lightweight attention architecture and bidirectional feature digging capture long-term dependencies in sequences, reducing the temporal forgetting problem of GRU structures.}

Fig. \ref{fig:fig3} demonstrates the use of text backward hidden layer features to mine text forward hidden layer features. According to the Equation \ref{eqnGA}, we can easily get the attention features $\tilde{\bm{\mathcal{H}}}^f$ to enhance global connectivity,
\begin{equation}
\tilde{\bm{\mathcal{H}}}^f = \bm{\mathcal{G}}(\bm{\mathcal{H}}^f),
\end{equation}
and fused with $\bm{\mathcal{H}}^f$ to obtain the forward joint features with one-way inference and global correlation as
\begin{equation}
\bm{\mathcal{T}}_f = \tilde{\bm{\mathcal{H}}}^f + \bm{\mathcal{H}}^f.
\end{equation}

To mine key information in the backward hidden layer $\bm{\mathcal{H}}^b$ , the backward probability matrix is obtained by global augmentation and nonlinear mapping as
\begin{equation}
\bm{\mathcal{P}}^b = \bm{\mathcal{F}}( \bm{\mathcal{G}}(\bm{\mathcal{H}}^b) ),
\end{equation}
where $\bm{\mathcal{F}}(\cdot)$ is 2-layer fully connected layer,and the interactive features by combining forward joint features with backward probability matrix by dot product operation as
\begin{equation}
\bm{\mathcal{T}}_{f\odot b}=\bm{\mathcal{T}}_f \odot \bm{\mathcal{P}}^b,
\end{equation}
where $\bm{\mathcal{T}}_{f\odot b}$ denote interactive features by using text backward hidden layer features to mine text forward hidden layer features.

According to the same calculation we can get $\bm{\mathcal{T}}_{b\odot f}$. For a joint representation, we add them, element by element, to get $\bm{\mathcal{T}}_c = \bm{\mathcal{T}}_{f\odot b}+\bm{\mathcal{T}}_{b\odot f}$ and decode them to get the word-level textual features $\bm{F}_G \in \mathbb{R}^{N_c \times d}$, i.e.,
\begin{equation}
\bm{F}_G =M L P\left(\bm{\mathcal{T}}_c\right) + \bm{\mathcal{T}}_c.
\end{equation}
\subsection{Regional-Oriented Attention Module}\label{sectionAttention}
To explore the intrinsic connection between vision and language, the ROAM module adaptively adjusts the distance between the final visual and textual embeddings in the embedding space by guiding the representation of multiscale visual features and and word-level textual features with regional visual features. It has two parts: 1) Intra-modal Fusion Attention (IFA) fuses regional visual features and multiscale visual features, and 2) Inter-modal Guidance Attention (IGA) employs regional visual features to guide textual features. Like most attention mechanism \cite{bahdanau2014neural} methods, the IFA and IGA have encoding and decoding parts, as shown in Fig. \ref{fig:fig4}.

\subsubsection{Intra-modal Fusion Attention (IFA)}
\begin{figure}[t]
  \centering
  \includegraphics[width=\linewidth]{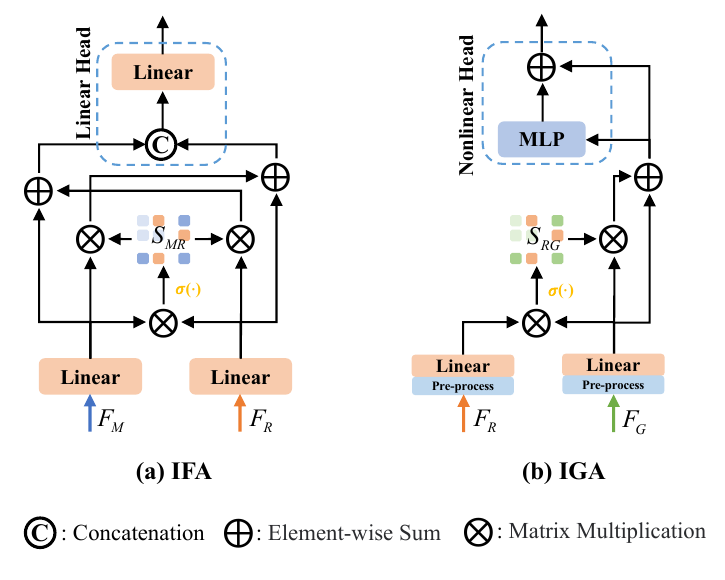}
  \caption{ROAM module: (a) IFA module; (b) IGA module.}
  \label{fig:fig4}
\end{figure}

To find fine-grained visual representation \textcolor{black}{\cite{9471793}} while solving the visual-semantic imbalance problem, we propose IFA module to find the commonality between multiscale and regional visual features. We fuse them to obtain an optimal visual representation pattern, as shown in Fig. \ref{fig:fig4}(a). We transform by linear as
\begin{equation}
\bm{F}_M^{\prime}= \bm{F}_M \bm{W}_M+\bm{b}_M,
\end{equation}
\begin{equation}
\bm{F}_R^{\prime}= \bm{F}_R \bm{W}_R+\bm{b}_R,
\end{equation}
where $\bm{W}_M (\bm{W}_R)$ and $\bm{b}_M (\bm{b}_R)$ are the respective weights and bias of a fully-connected layer. After matrix multiplication, we calculate the score of the two features, and use it to activate the two input features separately. Finally, we can obtain the converged features activated by the other modality as
\begin{equation}
\bm{S}_{M R} =\sigma \left(\bm{F}_M^{\prime} \bm{F}_{R}^{\prime \mathrm{T}}\right),
\end{equation}
\begin{equation}
\bm{F}_{R \oplus M} =\bm{S}_{M R} \bm{F}_R^{\prime}+\bm{F}_M^{\prime},
\end{equation}
\begin{equation}
\bm{F}_{M \oplus R} = \bm{S}_{M R}^{\mathrm{T}} \bm{F}_M^{\prime}+\bm{F}_R^{\prime},
\end{equation}
where $\bm{S}_{M R}$ denotes the input matrix multiplication result, and $\bm{F}_{R \oplus M}$ and $\bm{F}_{M \oplus R}$ are the converged features. We activate the features of visual modality using a linear head to get the fused features $\bm{F}_{M R} \in \mathbb{R}^{(N_m+N_r) \times d}$ as
\begin{equation}
\bm{F}_{M R}= Concat\left(\bm{F}_{R \oplus M} , \bm{F}_{M \oplus R }\right) \bm{W}_L+\bm{b}_L,
\end{equation}
where $\bm{W}_L$ and $\bm{b}_L$ respectively represent the weights and bias, and $Concat(\cdot)$ denotes concatenation operation.
\subsubsection{Inter-modal Guidance Attention (IGA)}
For better alignment with visual embedding while minimizing the distance between visual and textual embeddings in the latent embedding space, IGA module guide textual feature representation using regional visual features, as shown in Fig. \ref{fig:fig4}(b). Different texts have different lengths, which cannot directly interact with the fixed-length image features, so interaction between different modalities must be pre-processed. We let $\bm{F}_R = [\bm{r}_1,\bm{r}_2,...,\bm{r}_{N_r}]^{\mathrm{T}} \in \mathbb{R}^{N_r \times d}$ and $\bm{F}_G = [\bm{g}_1,\bm{g}_2,...,\bm{g}_{N_c}]^{\mathrm{T}} \in \mathbb{R}^{N_c \times d}$, and calculate the average values as
\begin{equation}
\bm{E}_R=\frac{1}{N_r} \sum_{i=1}^{N_r} \bm{r}_i , \ \bm{E}_G= \frac{1}{N_c} \sum_{i=1}^{N_c} \bm{g}_i,
\end{equation}
and we respectively expand $\bm{E}_R \in \mathbb{R}^{d}$ and $\bm{E}_G \in \mathbb{R}^{d}$ to obtain $\bm{E}_R^{\prime} \in \mathbb{R}^{B_c \times d}$ and $\bm{E}_G^{\prime} \in \mathbb{R}^{B_c \times d}$, where $B_c$ is the batch size of text input. We linearly transform these pre-processed features as
\begin{equation}
\bm{F}_R^{\prime}=\bm{E}_R^{\prime} \bm{W}_R +\bm{b}_R,
\end{equation}
\begin{equation}
\bm{F}_G^{\prime}=\bm{E}_G^{\prime} \bm{W}_G+\bm{b}_G,
\end{equation}
where $\bm{W}_R (\bm{W}_G)$ and $\bm{b}_R (\bm{b}_G)$ respectively are the weights and bias of a fully-connected layer. Similar treatments to the above, we calculate the score of input features, and use it to activate the textual features to obtain the regional-activated textual features $\bm{F}_{R \oplus G}$ as
\begin{equation}
\bm{S}_{R G} =\sigma\left(\bm{F}_R^{\prime} \bm{F}_G^{\prime \mathrm{T}}\right),
\end{equation}
\begin{equation}
\bm{F}_{R \oplus G} =\bm{S}_{R G} \bm{F}_G^{\prime}+\bm{F}_G^{\prime},
\end{equation}
where $\bm{S}_{R G}$ represents the input matrix multiplication result, and $\bm{F}_{R \oplus G}$ denotes the regional-activated textual features. We activate the features using a nonlinear head, and get the the final textual features $\bm{F}_{R G} \in \mathbb{R}^{B_c \times d}$ as
\begin{equation}
\bm{F}_{R G}=M L P\left(\bm{F}_{R \oplus G}\right)+\bm{F}_{R \oplus G}.
\end{equation}

\subsection{Objective Function}
To obtain the final visual and textual embeddings, we let $\bm{F}_M = [\bm{m}_1,\bm{m}_2,...,\bm{m}_{N_m}]^{\mathrm{T}} \in \mathbb{R}^{N_m \times d}$ and calculate the average values as
\begin{equation}
\bm{V}_{M}=\frac{1}{N_m} \sum_{i=1}^{N_m} \bm{m}_i.
\end{equation}
Similarly, we can respectively obtain $\bm{T}_G$, $\bm{V}_{M R}$, and $\bm{T}_{R G}$ from $\bm{F}_G$, $\bm{F}_{M R}$, and $\bm{F}_{R G}$. We follow \cite{yuan2022exploring,yuan2022remote} to employ the bidirectional triplet ranking loss \cite{karpathy2015deep},
\begin{equation}\label{eqn21}
\begin{aligned}
\mathcal{L}(V, T)=\sum_{\hat{T}} [\alpha-S(V, T)&+S(V, \hat{T})]_{+} \\
&+\sum_{\hat{V}}[\alpha-S(V, T)+S(\hat{V}, T)]_{+},
\end{aligned}
\end{equation}
where $\alpha$ is a margin parameter, $[x]_{+} \equiv \max (x, 0)$, $\hat{V}$ and $\hat{T}$ \textcolor{black}{are images and text correspond to negative samples} in the mini-batch, $S\left(\cdot, \cdot \right)$ is the Cosine function.

To ensure that the original visual and textual semantics remain unchanged, we add a global visual-semantic constraint to serve as an external constraint for the final visual and textual representations. We combine the two triplet losses to obtain the total objective function,
\begin{equation}\label{eqn22}
\mathcal{L}_{total} = \mathcal{L}(\bm{V}_{M R}, \bm{T}_{R G}) + \lambda_g \mathcal{L}(\bm{V}_M, \bm{T}_G),
\end{equation}
where $\mathcal{L}(\bm{V}_{M R}, \bm{T}_{R G})$ is the final ranking loss, $\mathcal{L}(\bm{V}_M, \bm{T}_G)$ is the global visual-semantic constraint, and $\lambda_g$ is the constraint parameter.
\begin{figure}[t]
  \centering
  \includegraphics[width=\linewidth]{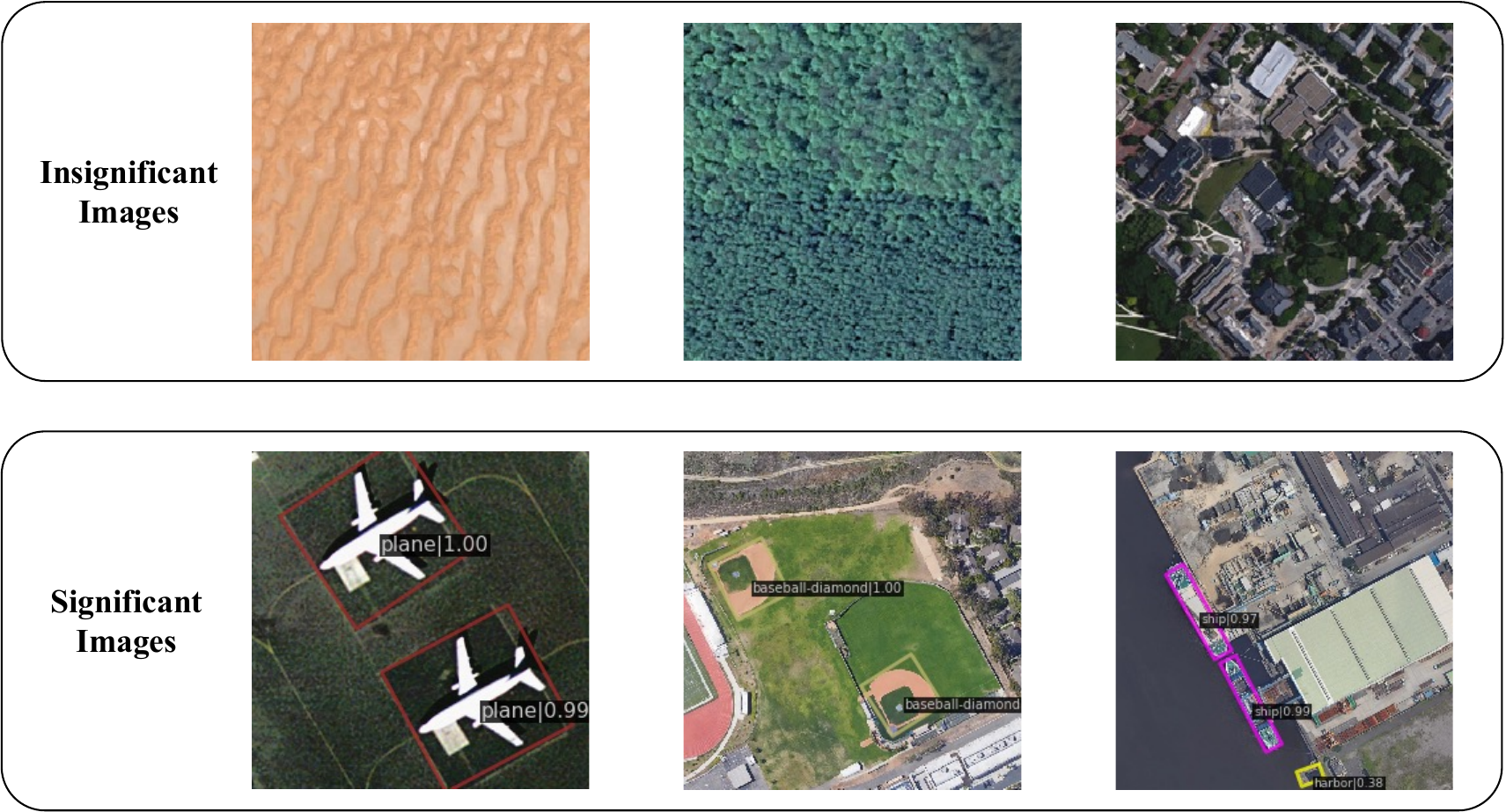}
  \caption{Insignificant and significant images divided by whether it contains significant objects or not.}
  \label{fig:fig10}
\end{figure}
\section{EXPERIMENTS}
\subsection{Datasets and Metrics}
\subsubsection{Datasets}
We evaluated our model on the RSICD and RSITMD datasets. RSICD \cite{lu2017exploring} contains 10,921 images, each image is associated with five sentences. We followed Yuan et al. \cite{yuan2022exploring, yuan2022remote} in dividing the dataset into 7,862 training images, 1,966 validation images, and 1,093 test images. RSITMD \cite{yuan2022exploring} contains 4,743 images and five sentences at a more ﬁne-grained description than RSICD. Using the same division as \cite{yuan2022exploring, yuan2022remote}, we obtained 3,435 training images, 856 validation images, and 452 test images. In comparison experiments, we randomly disrupted the training and validation sets, and used cross-validation for three experiments to find the final average experimental results. Specifically, we fix the training and validation sets by picking a certain random allocation result in the ablation experiments to reduce the impact of different data distributions on model performance.

To explore the performance of our model on significant and insignificant sample pairs, we divided the RSICD and RSITMD test sets. We adopt the following division: firstly, we divide the test sets into significant and insignificant sample pairs 
\textcolor{black}{according to whether the image contains categories of common remote sensing objects or not, using RoI encoder \footnote{\textcolor{black}{https://github.com/open-mmlab/mmrotate}} pretrained on the remote sensing general object detection dataset DOTA-v2.0 \cite{xia2018dota} with 18 object categories \footnote{\textcolor{black}{18 object categories include: plane, ship, storage tank, baseball diamond, tennis court, basketball court, ground track field, harbor, bridge, large vehicle, small vehicle, helicopter, roundabout, soccer ball field, swimming pool, container crane, airport and helipad.}} to detect the objects.} \textcolor{black}{To prevent some misidentification, }we manually screen these test sets \textcolor{black}{to identify whether the object category could be clearly recognized by the human eye}. The final division results (partially) are shown in Fig. \ref{fig:fig10}.
\subsubsection{Metrics}
As with most image-text retrieval methods, we evaluated the performance of the retrieval algorithm by $R@K (K=1, 5, 10)$ and $mR$, where $R@K$ is the percentage of correctly matched pairs among the top $K$ retrieval results, and $mR$ is the average values of $R@K$.

\begin{figure}[t]
  \centering
  \includegraphics[width=0.75\linewidth]{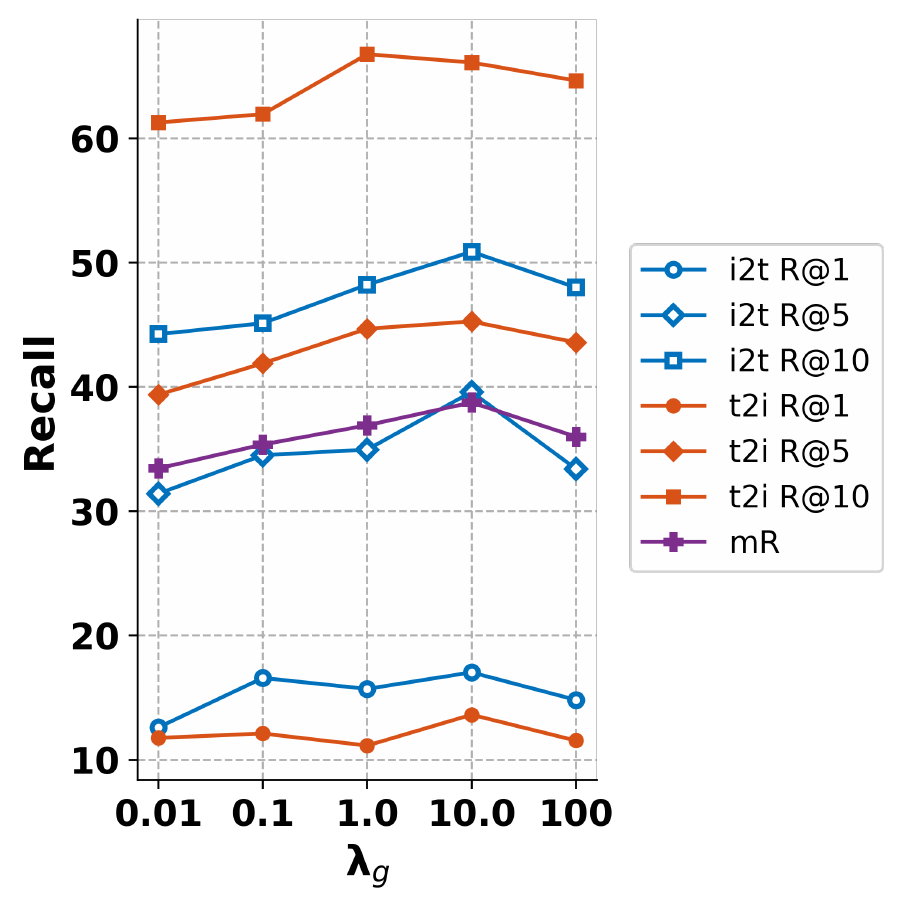}
  \caption{Results of sentence retrieval (i2t) and image retrieval (t2i) at different values of $\lambda_g$. As $\lambda_g$ increases, the global visual-semantic constraint effect is enhanced, and the regional visual feature dependency is diminished.}
  \label{fig:fig5}
\end{figure}
\subsection{Implementation Details}
All experiments were conducted in a station equipped with NVIDIA RTX A6000. To ensure that the experiment was reproducible, and to reduce the effect of random factors, we set a fixed random seed the same as \cite{rao2022does}. We used Adam \cite{kingma2014adam} as the model optimizer, and set the initial learning rate to 0.0002 with decays of 0.7 every 20 epochs. The mini-batch size is set 100, and the embedding size is set as 512. We set the epoch to 50 on both datasets. In the GA module, there are 2 headers for text feature encoding. We froze the pretrained ResNet, and the margin $\alpha$ in Equation \ref{eqn21} was set at 0.2. After experimental verification, we set $\lambda_g$ to 10.0 in Equation \ref{eqn22}. The model that performed best in the validation is set for testing, and used for the average results.

\begin{table*}[htbp]
\renewcommand\arraystretch{1.5}
\caption{Comparisons of image-text retrieval results on RSICD and RSITMD.}
\label{tab:table1}
\resizebox{\linewidth}{!}{
\begin{tabular}{cccccccc}
\hline
\multicolumn{2}{c}{} & \multicolumn{3}{c}{RSICD   dataset} & \multicolumn{3}{c}{RSITMD dataset} \\ \hline
\textbf{Method} & \multicolumn{1}{c|}{\textcolor{black}{\textbf{Parms}}} & \textbf{\begin{tabular}[c]{@{}c@{}}Image-query-Text  \\ R@1 / R@5 / R@10\end{tabular}} & \textbf{\begin{tabular}[c]{@{}c@{}}Text-query-Image   \\ R@1 / R@5 / R@10\end{tabular}} & \multicolumn{1}{c|}{\textbf{mR}} & \textbf{\begin{tabular}[c]{@{}c@{}}Image-query-Text\\ R@1 / R@5 / R@10\end{tabular}} & \textbf{\begin{tabular}[c]{@{}c@{}}Text-query-Image\\ R@1 / R@5 / R@10\end{tabular}} & \textbf{mR} \\ \hline
\multicolumn{8}{c}{Traditional Image-Text Retrieval Methods} \\ \hline
VSE++ & \multicolumn{1}{c|}{29 M} & 4.56 / 16.73 / 22.94 & 4.37 / 15.37 / 25.35 & \multicolumn{1}{c|}{14.89} & 9.07 / 21.61 / 31.78 & 7.73 / 27.80 / 41.00 & 23.17 \\
SCAN i2t & \multicolumn{1}{c|}{60 M} & 4.82 / 13.66 / 21.99 & 3.93 / 15.20 / 25.53 & \multicolumn{1}{c|}{14.19} & 8.92 / 22.12 / 33.78 & 7.43 / 25.71 / 39.03 & 22.83 \\
SCAN t2i & \multicolumn{1}{c|}{60 M} & 4.79 / 16.19 / 24.86 & 3.82 / 15.70 / 28.28 & \multicolumn{1}{c|}{15.61} & 7.01 / 20.58 / 30.90 & 7.06 / 26.49 / 42.21 & 22.37 \\
CAMP & \multicolumn{1}{c|}{63 M} & 4.64 / 14.61 / 24.09 & 4.25 / 15.82 / 27.82 & \multicolumn{1}{c|}{15.20} & 8.11 / 23.67 / 34.07 & 6.24 / 26.37 / 42.37 & 23.47 \\
CAMERA & \multicolumn{1}{c|}{64 M} & 4.57 / 13.08 / 21.77 & 4.00 / 15.93 / 26.97 & \multicolumn{1}{c|}{14.39} & 8.33 / 21.83 / 33.11 & 7.52 / 26.19 / 40.72 & 22.95 \\ \hline
\multicolumn{8}{c}{Remote Sensing Image-Text Retrieval Methods} \\ \hline
LW-MCR & \multicolumn{1}{c|}{-} & 3.29 / 12.52 / 19.93 & 4.66 / 17.51 / 30.02 & \multicolumn{1}{c|}{14.66} & 10.18 / 28.98 / 39.82 & 7.79 / 30.18 / 49.78 & 27.79 \\
AMFMN & \multicolumn{1}{c|}{36 M} & 5.21 / 14.72 / 21.57 & 4.08 / 17.00 / 30.60 & \multicolumn{1}{c|}{15.53} & 10.63 / 24.78 / 41.81 & 11.51 / 34.69 / 54.87 & 29.72 \\
GaLR & \multicolumn{1}{c|}{46 M} & 6.59 / 19.85 / 31.04 & 4.69 / 19.48 / 32.13 & \multicolumn{1}{c|}{18.96} & 14.82 / 31.64 / 42.48 & 11.15 / 36.68 / 51.68 & 31.41 \\
KCR & \multicolumn{1}{c|}{-} & 5.95 / 18.59 / 29.58 & 5.40 / \underline{22.44} / 37.36 & \multicolumn{1}{c|}{19.89} & - & - & - \\
SWAN & \multicolumn{1}{c|}{40 M} & 7.41   / 20.13 / 30.86 & 5.56   / 22.26 / \underline{37.41} & \multicolumn{1}{c|}{\underline{20.61}} & 13.35 / \underline{32.15} / \underline{46.90} & 11.24 / 40.40 / \underline{60.60} & 34.11 \\
HVSA & \multicolumn{1}{c|}{-} & 7.47   / \underline{20.62} / \underline{32.11} & 5.51   / 21.13 / 34.13 & \multicolumn{1}{c|}{20.16} & 13.20 / 32.08 / 45.58 & 11.43 / 39.20 / 57.45 & 33.16 \\ \hline
\textcolor{black}{\textbf{DOVE-S}} & \multicolumn{1}{c|}{25 M} & \underline{7.62}   / 19.85 / 30.65 & \underline{5.84}   / 21.87 / 36.54 & \multicolumn{1}{c|}{20.40} & \underline{15.71} / 31.86 / 44.69 & \underline{12.09} / \underline{42.39} / 59.16 & \underline{34.32} \\
\textbf{DOVE} & \multicolumn{1}{c|}{38 M} & \textbf{8.66   / 22.35 / 34.95} & \textbf{6.04   / 23.95 / 40.35} & \multicolumn{1}{c|}{\textbf{22.72}} & \textbf{16.81 / 36.80 / 49.93} & \textbf{12.20 / 44.13 / 66.50} & \textbf{37.73} \\ \hline
\end{tabular}}
\end{table*}

\begin{table}[t]
\renewcommand\arraystretch{1.5}
\caption{Comparison experiments of different DTGA input combinations on RSITMD test set.}
\label{tab:table2}
\resizebox{\linewidth}{!}{
\begin{tabular}{cccccccc}
\hline
\multirow{2}{*}{\textbf{DTGA Input}}                                            & Sentence Retrieval      & Image Retrieval               &                \\ \cline{2-3}
 & \textbf{R@1 / R@5 / R@10}      & \textbf{R@1 / R@5 / R@10}               & \textbf{mR}             \\ \hline
\multicolumn{1}{c|}{$\bm{\mathcal{H}}^f$, $\bm{\mathcal{H}}^f$}                 & 15.93 / 33.19 / 46.68 & 13.50 / 44.20 / 64.51          & 36.33          \\
\multicolumn{1}{c|}{$\bm{\mathcal{H}}^b$, $\bm{\mathcal{H}}^b$}                 & 15.27 / 36.06 / 50.44 & 13.89 / 44.82 / 66.11         & 37.77         \\
\multicolumn{1}{c|}{$\bm{\mathcal{H}}^f$, $\bm{\mathcal{H}}^b$}                 & \textbf{17.04} / \textbf{39.60} / \textbf{50.88} & \textbf{13.63} / \textbf{45.27} / \textbf{66.11} & \textbf{38.75} \\
\multicolumn{1}{c|}{$\bm{\mathcal{H}}^{\frac{f+b}{2}}$, $\bm{\mathcal{H}}^{\frac{f+b}{2}}$} & 15.49 / 36.06 / 50.44 & 13.10 / 43.27 / 65.49          & 37.31         \\ \hline
\end{tabular}}
\end{table}
\subsection{Parameter Evaluation}
\subsubsection{Evaluating the Impact of global visual-semantic constraint}
To explore the impact of the global visual-semantic constraint on retrieval performance, we set up a set of experiments. Fig. \ref{fig:fig5} shows the retrieval performance under different values of $\lambda_g$. As $\lambda_g$ increases, the global visual-semantic constraint effect is enhanced, and the regional visual feature dependency is diminished. We can also find that $mR$ is lowest when $\lambda_g=0.01$. As $\lambda_g$ increases, the retrieval performance increases, and the overall retrieval performance reaches the maximum when $\lambda_g=10.0$. Overall retrieval performance begins to decline after $\lambda_g=10.0$, indicating that excessive global semantic constraints can have an antagonistic effect. In particular, t2i $R@10$ reaches its highest at $\lambda_g=1.0$, while it reaches second place at $\lambda_g=10.0$. The above results indicate that the global visual-semantic constraints are enhanced, and somewhat reduced involvement in regional-oriented embeddings can improve retrieval performance.

\begin{figure*}[t]
  \centering
  \includegraphics[width=\linewidth]{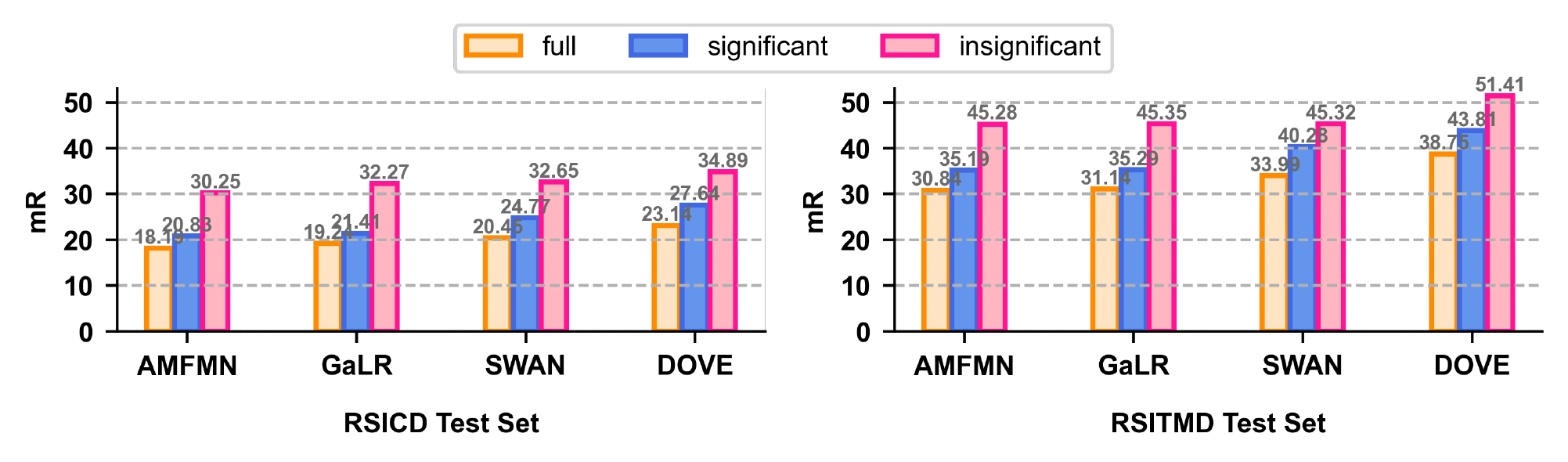}
  \caption{Results on the full, significant and insignificant test sets of the RSICD and RSITMD datasets \textcolor{black}{for testing the recognition of different retrieval methods on significant and insignificant objects. The \textit{full} dataset is divided into \textit{significant} and \textit{insignificant} parts according to whether the image contains categories of common remote sensing objects or not.}}
  \label{fig:fig11}
\end{figure*}

\subsubsection{Evaluation of DTGA Module}
We set up experiments on RSITMD dataset with different combinations of DTGA inputs: 1) $\bm{\mathcal{H}}^f$, $\bm{\mathcal{H}}^f$ referring to both inputs are the output of the forward hidden layer of the GRU, 2) $\bm{\mathcal{H}}^b$, $\bm{\mathcal{H}}^b$ referring to both inputs are the output of the backward hidden layer of the GRU, 3) $\bm{\mathcal{H}}^f$, $\bm{\mathcal{H}}^b$ referring to the two inputs respectively use the forward and backward hidden layer outputs of the GRU, and 4) $\bm{\mathcal{H}}^{\frac{f+b}{2}}$, $\bm{\mathcal{H}}^{\frac{f+b}{2}}$ referring to both inputs are the average of the forward and backward hidden layer of the GRU. The above experimental results are presented in Table \ref{tab:table2}. Compare the input combination is $\bm{\mathcal{H}}^f$, $\bm{\mathcal{H}}^f$ and $\bm{\mathcal{H}}^b$, $\bm{\mathcal{H}}^b$, there are higher $R@1$ and lower $R@10$ when the input is the former, and lower $R@1$ and higher $R@10$ when the input is the latter. We found the overall retrieval performance to be highest when the input combination  is $\bm{\mathcal{H}}^f$, $\bm{\mathcal{H}}^b$. The above experiments show that DTGA input combinations have different effects on retrieval performance. It also demonstrated that using the input combination is $\bm{\mathcal{H}}^f$, $\bm{\mathcal{H}}^b$ can improve overall retrieval performance, which can enhance the semantic representation of the text.

\subsection{Quantitative Comparison}
We compared the DOVE model with traditional image-text retrieval methods VSE++, SCAN, CAMP, and CAMERA, and current mainstream remote sensing image-text retrieval methods LW-MCR, AMFMN, GaLR, and KCR, on the RSICD and RSITMD datasets. As the original paper of traditional image-text retrieval methods do not conduct experiments on RSICD and RSITMD, to make fair comparisons, we used the same image encoder and text encoder as our model, conductd three experiments, and averaged the results for the traditional image-text retrieval methods. Furthermore, we directly quote the original paper's best results for the remote sensing image-text retrieval methods. \textcolor{black}{To reduce the impact of parameter complexity, we added a small-size DOVE model, DOVE-S, using ResNet-18 \cite{he2016deep} as the MSV encoder.}
\begin{itemize}
\item VSE++ \cite{faghri2017vse++} embeds the full image and sentence into an embedding space and calculates their similarity;
\item SCAN \cite{lee2018stacked} aligns regional visual features and word-level textual features using attention mechanisms;
\item CAMP \cite{wang2019camp} explores the intrinsic connection between images and text through cross-modal interaction;
\item CAMERA \cite{qu2020context} summarizes region-level representation from multiple views to achieve cross-modal semantic alignment;
\item LW-MCR \cite{yuan2021lightweight} uses group convolution and visual attention for a lightweight image-text retrieval model;
\item AMFMN \cite{yuan2022exploring} uses multiscale visual features to guide textual representation and dynamically filter redundant features;
\item GaLR \cite{yuan2022remote} dynamically fuses global and local visual features to improve visual representation;
\item KCR \cite{mi2022knowledge} enriches text semantics to improve textual representation by introducing an external knowledge graph;
\item SWAN \cite{pan2023reducing} reduces the semantic confusion zones in the embedding space to improve the fine-grained perception of the scene.
\item HVSA \cite{zhang2023hypersphere} solves the characteristics of data distribution and the varying difficulty levels of different sample pairs via curriculum learning.
\end{itemize}

\subsubsection{Results on RSICD Dataset}
Table \ref{tab:table1} shows the experimental results on RSICD, from which we can find that the DOVE model shows a significant increase compared with state-of-the-art methods. For example, $R@1$ improved by 16.9\% (8.66 vs. 7.41)and 8.6\% (6.04 vs. 5.56) in sentence and image retrieval, respectively. In total, there is a 10.2\% (22.72 vs. 20.61) improvement on the $mR$ metric. \textcolor{black}{The results of DOVE-S experiments indicate that the small-size DOVE model outperforms LW-MCR \cite{yuan2021lightweight}, AMFMN \cite{yuan2022exploring}, and GaLR \cite{yuan2022remote} using ResNet-18 \cite{he2016deep} as the backbone while being close to that of the SOTA methods and providing a relatively significant improvement in R@1. So we can say that our method outperformed traditional and remote sensing retrieval methods on RSICD dataset, reflecting it superior performance at solving visual-semantic imbalance.}
\subsubsection{Results on RSITMD Dataset}
RSITMD has a more fine-grained sentence description than RSICD; hence its overall retrieval performance can be improved. Table \ref{tab:table1} shows the performance on the RSITMD test set; our method has improved in almost all metrics. Depending on the superior of visual representation, the value of $R@10$ for image retrieval reached 66.50 on RSITMD test set, which is a 9.7\% (66.50 vs. 60.60) improvement, and in general, there is a 10.6\% (37.73 vs. 34.11) improvement on the $mR$ metric. \textcolor{black}{The DOVE-S results show that small-size DOVE significantly improves R@1 and can outperform the SOTA methods on mR.} To sum up, it can be seen that our model has a high performance advantage, because it has a substantial improvement in most of the metrics compared to the state-of-the-art methods.
\subsubsection{Results on significant and insignificant test sets}
In order to further explore the algorithm's ability to match significance and insignificance, we set up several sets of experiments with controls: 1) \textbf{full} for the full test set; 2) \textbf{significant} for the significant test set, which is used to test the model's significance matching ability; and 3) \textbf{insignificant} for the insignificant test set, which is used to test the model's insignificance matching ability. Fig. \ref{fig:fig11} shows results on the full, significant and insignificant test sets of the RSICD and RSITMD datasets. \textcolor{black}{The evaluation of visual significance in images is nuanced, and retrieval performance in remote sensing datasets is influenced by visual and textual elements, where simplistic text can lead to ambiguous semantics. The mR metric reflects the average level of ranking of matching images or text in the current dataset. The significant and insignificant test set images interacted with each other in ranking, causing the mR in the full test set to decrease.} It is observed that the current algorithm performs better for the insignificant test set and not so well for the significant test set, while the overall retrieval performance is the lowest. It is clear that the retrieval algorithm matches more significant image-text pairs and less well on insignificant ones, resulting in an overall retrieval performance is lower than the significant and insignificant test set performance. In the significance matching results, comparing the best competitor SWAN, the retrieval performance is improved by 11.6\% (24.77 vs. 27.64) on the RSICD significant test set and by 8.8\% (40.28 vs. 43.81) on the RSITMD significant test set. For insignificant matching results, comparing the best competitor SWAN, retrieval performance is improved by 6.9\% (32.65 vs. 34.89) on the RSICD insignificant test set, and by 13.4\% (45.32 vs. 51.41) on the RSITMD insignificant test set. Our method clearly improves the ability to match insignificant sample pairs significantly, which further improves the overall retrieval performance.
\begin{figure}[t]
  \centering
  \includegraphics[width=\linewidth]{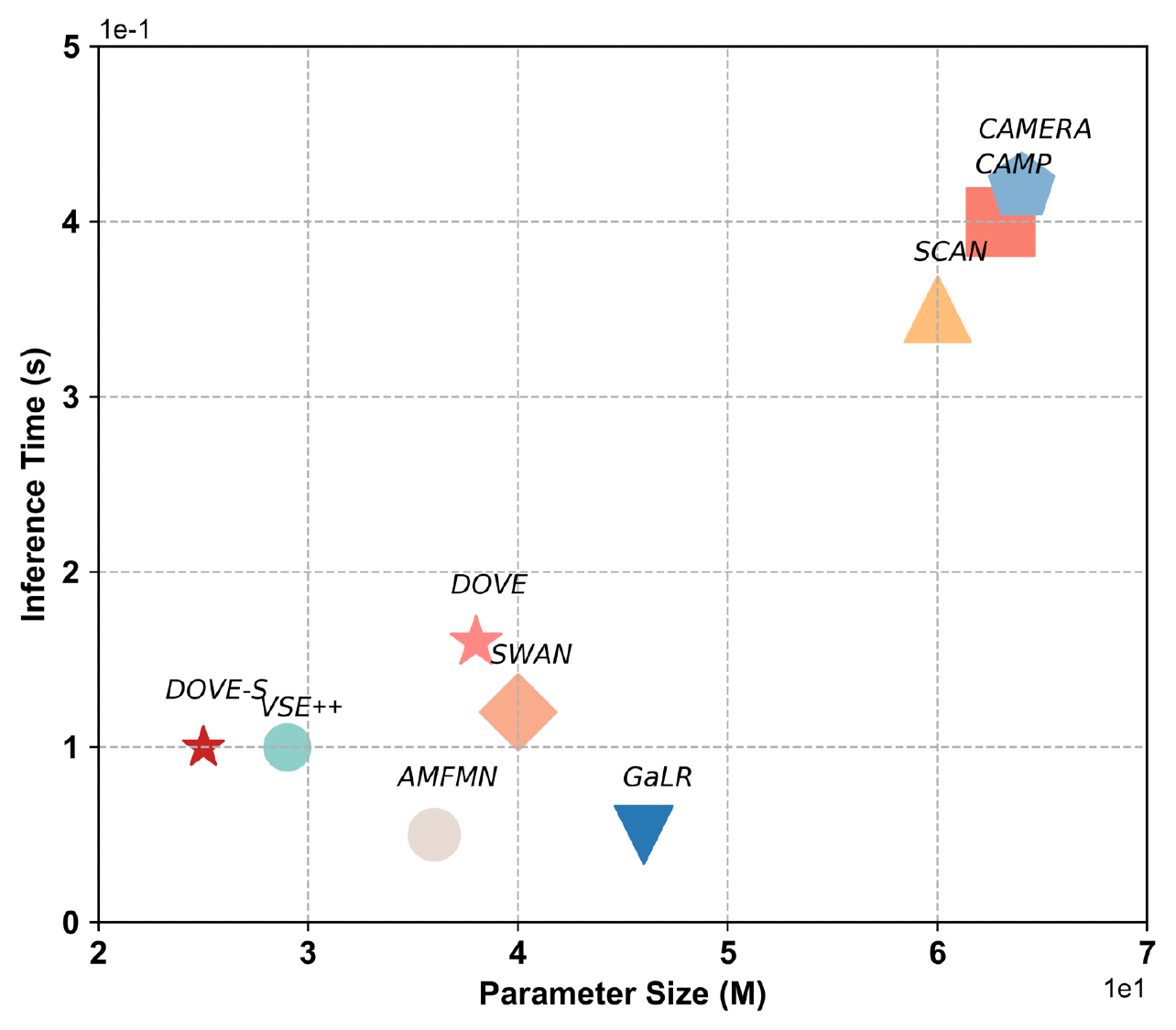}
  \caption{\textcolor{black}{Parameter size and inference time of remote sensing image-text retrieval methods. Different colored shapes represent different methods, where the size of the shape indicates the size of the parameter.}}
  \label{fig:fig12}
\end{figure}

\subsubsection{\textcolor{black}{Comparison of average inference time}}
    \textcolor{black}{To explore the performance of different remote sensing image-text retrieval methods on time consumption, we comprehensively considered parameter size and average inference time  to these methods as shown in Fig. \ref{fig:fig12}. Traditional image-text retrieval methods such as SCAN, CAMP, and CAMERA have higher inference times with large parameters. In comparison, remote sensing image-text retrieval methods have lower inference times with smaller parameters. Our proposed DOVE method has a shorter inference time with smaller parameters, whereas DOVE-S achieves the lowest inference time with the most minor parameters. Combined with the above performance comparisons in Table \ref{tab:table1}, our DOVE approach can achieve high performance with only a few parameters and low time consumption.}

\begin{table}[t]
\renewcommand\arraystretch{1.5}
\caption{Ablation experiments on RSITMD test set.}
\label{tab:table3}
\resizebox{\linewidth}{!}{
\begin{tabular}{cccccccc}
\hline
\multicolumn{1}{c}{\multirow{2}{*}{\textbf{Method}}} & Sentence Retrieval        & Image Retrieval      &       \\ \cline{2-3} 
\multicolumn{1}{c}{}                                 & \textbf{R@1 / R@5 / R@10}        & \textbf{R@1 / R@5 / R@10}      & \textbf{mR}             \\ \hline
\multicolumn{1}{l|}{w/o DTGA}                        & 13.72 / 35.40 / 49.78            & 12.21 / 43.81 / 65.13          & 36.67          \\
\multicolumn{1}{l|}{w/o IFA}                         & 15.93 / 32.74 / 49.34            & 12.26 / 44.69 / 66.42          & 36.90         \\
\multicolumn{1}{l|}{w/o IGA}                         & 14.38 / 33.41 / 49.56 & 12.70 / 43.45 / 63.45         & 36.16          \\
\multicolumn{1}{l|}{full}                            & \textbf{17.04} / \textbf{39.60} / \textbf{50.88}   & \textbf{13.63} / \textbf{45.27} / \textbf{66.11} & \textbf{38.75} 
 \\ \hline
\end{tabular}}
\end{table}
\begin{table}[t]
\renewcommand\arraystretch{1.5}
\caption{Comparison experiments with different IFA/IGA Heads on RSITMD test set.}
\label{tab:table4}
\resizebox{\linewidth}{!}{
\begin{tabular}{cccccccc}
\hline
\textbf{IFA}  & \textbf{IGA}                    & Sentence Retrieval               & Image Retrieval               &                \\ \cline{3-4}
\textbf{Head} & \textbf{Head}                   & \textbf{R@1 / R@5 / R@10}      & \textbf{R@1 / R@5 / R@10}      & \textbf{mR}    \\ \hline
\textbf{L}    & \multicolumn{1}{c|}{\textbf{L}} & 14.16 / 34.96 / 49.56          & 11.86 / 41.86 / 65.00          & 36.23          \\
\textbf{L}    & \multicolumn{1}{c|}{\textbf{N}} & \textbf{17.04} / \textbf{39.60} / \textbf{50.88}          & \textbf{13.63} / \textbf{45.27} / \textbf{66.11} & \textbf{38.75} \\
\textbf{N}    & \multicolumn{1}{c|}{\textbf{L}} & 14.60 / 37.39 / 52.21
& 12.21 / 44.47 / 65.97          & 37.81          \\
\textbf{N}    & \multicolumn{1}{c|}{\textbf{N}} & 14.38 / 34.29 / 50.22          & 12.43 / 42.61 / 64.47          & 36.40          \\ \hline
\end{tabular}}
\end{table}
\subsection{Ablation Studies}
\subsubsection{Analysis of Ablation Experiments}
We conducted ablation experiments on RSITMD dataset (to verify the effect of different modules, as shown in Table \ref{tab:table3}. Because the ROAM module consists of IFA and IGA modules, we explore the roles of the IFA and IGA modules separately. We used three model blocks compared with the \textbf{full} experimental group: 1) \textbf{w/o DTGA} refers to the removal of the DTGA module, just using average values on the forward and backward hidden layer outputs of bidirectional GRU; 2) \textbf{w/o IFA} refers to the replacement of the original complex fusion method with regular concatenation; 3) \textbf{w/o IGA} refers to the removal of the IGA module. It can be found that the DTGA module can significantly improve retrieval performance, and sentence and image retrieval are increased by 24.2\% and 11.6\%, respectively, on the $R@1$ index. From w/o IFA, it can be found that image retrieval has improved by 11.2\% on the $R@1$ indicator. This shows that the IFA module can substantially improve image retrieval ability. Observing w/o IGA, we can find that the IGA module can also improve sentence and image retrieval performance. The above experimental results show that the proposed DOVE model can enhance the visual and textual representations and effectively solve the visual-semantic imbalance.

\subsubsection{Further Exploration of ROAM Module}
The decoding method will have different effects on intra-modal and inter-modal interactions. We set up experiments on RSITMD dataset to verify that different decoding methods affect the modality interaction, as shown in Table \ref{tab:table4}. \textbf{L} stands for \textbf{Linear Head}, which indicates linear decoding, and \textbf{N} stands for \textbf{Nonlinear Head}, which indicates nonlinear decoding, with the structure shown in Fig. \ref{fig:fig4}. The experimental comparison shows that the best overall retrieval performance is achieved when the IFA module uses linear decoding and the IGA module uses nonlinear decoding, whose image retrieval capability has improved by 19.6\% on $R@1$ metric compared to when the IFA module uses nonlinear decoding and the IGA module uses linear decoding. When both IFA and IGA modules use linear decoding or nonlinear decoding, there is a significant decrease in overall retrieval performance. The experimental results show that linear decoding for interactions between the same modality and nonlinear decoding for interactions between different modalities in image-text retrieval can improve retrieval performance. For homogeneous modality interactions, it is necessary to keep the original features unchanged as much as possible, while different modalities require further decoding to uncover deeper semantics.

\begin{figure}[t]
  \centering
  \includegraphics[width=\linewidth]{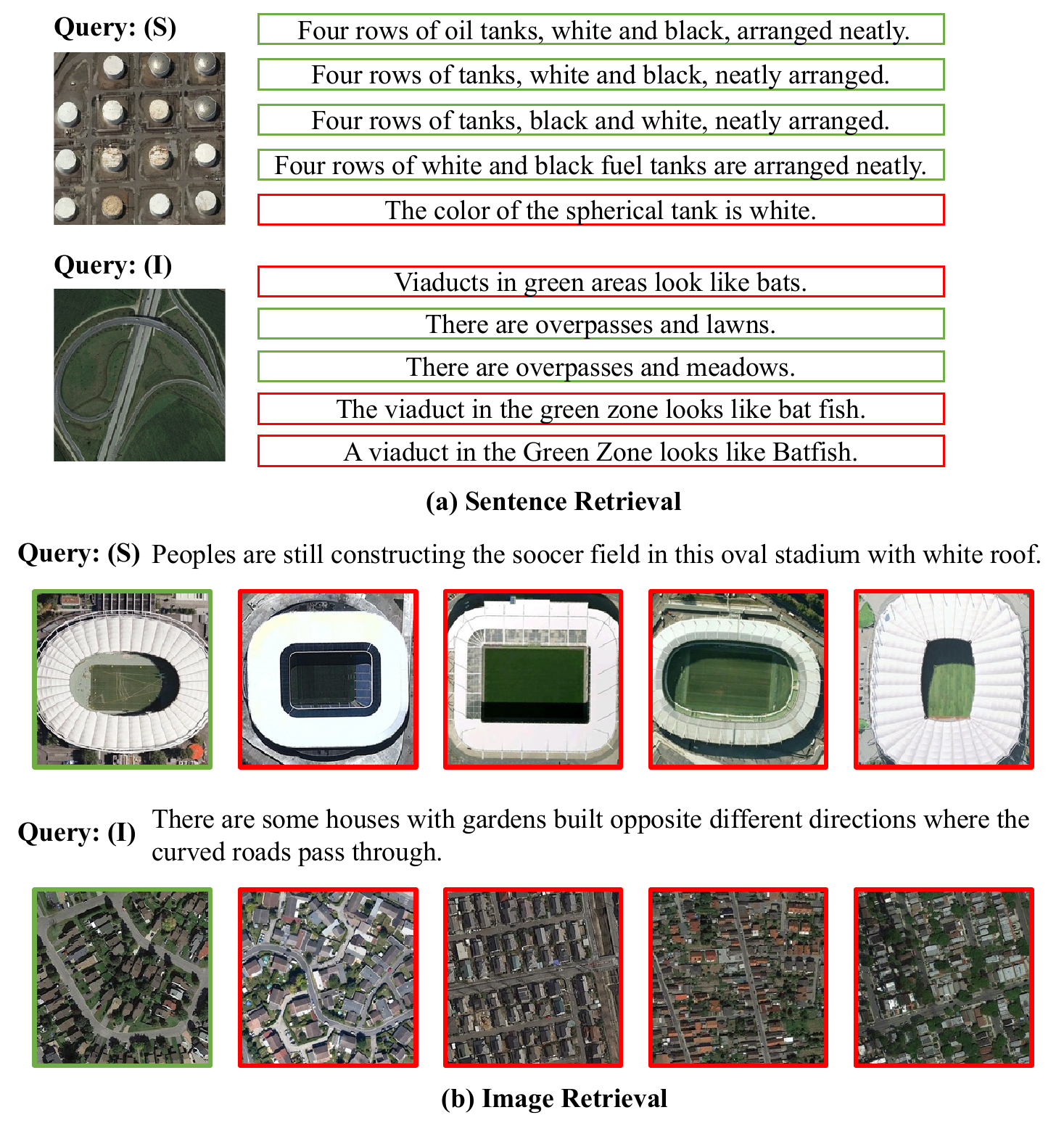}
  \caption{Qualitative results of bidirectional retrieval on RSITMD dataset: (a) Sentence Retrieval; (b) Image Retrieval. Green and red boxes indicate correct and incorrect matching results, respectively. (S) and (I) indicate significant and insignificant, i.e., whether the semantic object is significant or not.}
  \label{fig:fig6}
\end{figure}

\begin{figure}[t]
  \centering
  \includegraphics[width=\linewidth]{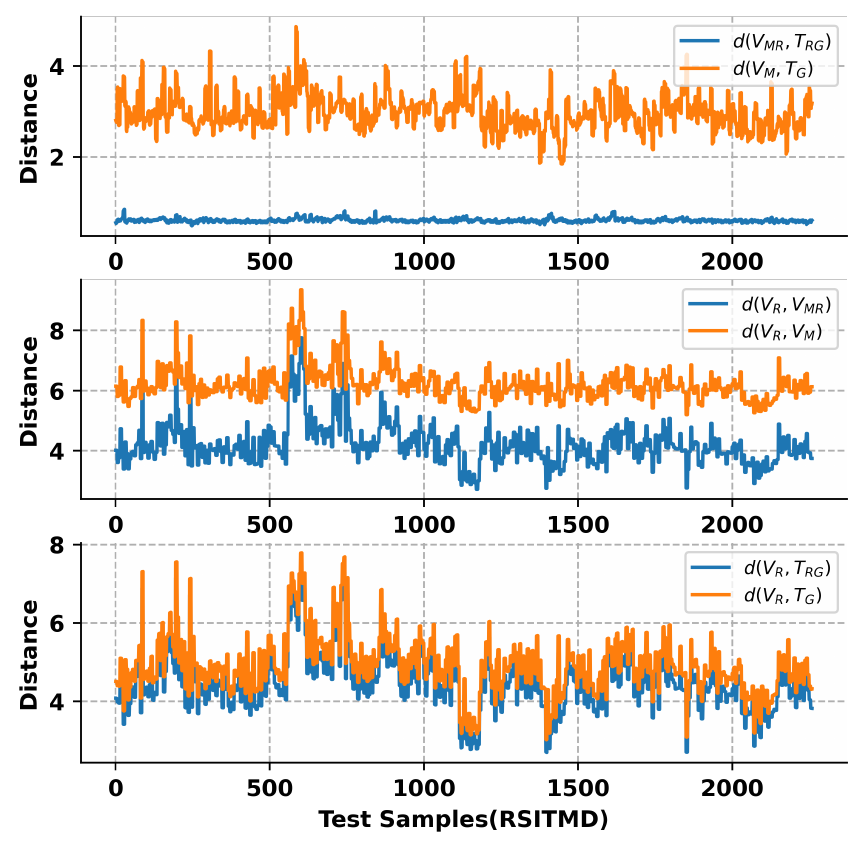}
  \caption{Distances between different embeddings on the RSITMD test set, where the test samples are all positive sample pairs.}
  \label{fig:fig9}
\end{figure}

\subsubsection{Exploring the Effects of Different Embedding Sizes and Mini-batch Sizes}
The retrieval model has different effects for different embedding sizes and mini-batch sizes. To further explore their effects, two groups of experiments on RSICD and RSITMD datasets investigated the effects of embedding and mini-batch sizes on our DOVE model separately, as shown in Table \ref{tab:table5}. In the experiment to explore the effect of embedding size, we set the mini-batch size to 100; in the experiment to explore the effect of mini-batch size, we set the embedding size to 512. Our DOVE model can show significant superiority for different embedding sizes. At embedding size 512, our model has the highest $mR$ on the RSICD and RSITMD datasets when the overall retrieval performance is the best. Observing the second group of experiments, we find that the optimal mini-batch size setting relates to the dataset size. For the RSICD dataset, the best performance is achieved with a mini-batch size of 128 orders of magnitude, while for the RSITMD dataset, the best performance is achieved with a mini-batch size of 32 or 64 orders of magnitude. Combining the above experiments, we find that our model has advantages in handling different embedding sizes, and setting the appropriate mini-batch size according to the size of the dataset can improve the model performance by a certain amount.

\begin{table*}[t]
\renewcommand\arraystretch{1.5}
\caption{Comparisons of different embedding sizes and mini-batch sizes on RSICD and RSITMD.}
\label{tab:table5}
\resizebox{\linewidth}{!}{
\begin{tabular}{cccccccc}
\hline
\multicolumn{2}{c}{} & \multicolumn{3}{c}{RSICD   dataset} & \multicolumn{3}{c}{RSITMD dataset} \\ \cline{3-8} 
\multicolumn{2}{c}{\multirow{-2}{*}{}} & Sentence Retrieval & Image Retrieval & \multicolumn{1}{c|}{} & Sentence Retrieval & Image Retrieval &  \\ \hline
\multicolumn{1}{c|}{\textbf{Type}} & \multicolumn{1}{c|}{\textbf{Size}} & \textbf{R@1 / R@5 / R@10} & \textbf{R@1 / R@5 / R@10} & \multicolumn{1}{c|}{\textbf{mR}} & \textbf{R@1 / R@5 / R@10} & \textbf{R@1 / R@5 / R@10} & \textbf{mR} \\ \hline
\multicolumn{1}{c|}{} & \multicolumn{1}{c|}{256} & {8.05 / 21.13 / 33.85} & {5.73 / 23.09 / 40.95} & \multicolumn{1}{c|}{{22.13}} & {14.60 / 34.07 / 49.34} & {11.77 / 44.73 / 67.39} & {36.98} \\
\multicolumn{1}{c|}{} & \multicolumn{1}{c|}{512} & {7.87 / 24.25 / 35.86} & {7.01 / 23.75 / 40.09} & \multicolumn{1}{c|}{{\textbf{23.14}}} & {17.04 / 39.60 / 50.88} & {13.63 / 45.27 / 66.11} & {\textbf{38.75}} \\
\multicolumn{1}{c|}{} & \multicolumn{1}{c|}{1024} & {7.04 / 20.31 / 32.20} & {6.55 / 23.22 / 39.67} & \multicolumn{1}{c|}{{21.50}} & {15.71 / 34.51 / 46.24} & {13.14 / 43.27 / 64.73} & {36.27} \\
\multicolumn{1}{c|}{\multirow{-4}{*}{\begin{tabular}[c]{@{}c@{}}Embedding \\ Size\end{tabular}}} & \multicolumn{1}{c|}{2048} & {8.14 / 21.59 / 33.21} & {5.95 / 22.69 / 39.34} & \multicolumn{1}{c|}{{21.82}} & {13.94 / 34.29 / 49.56} & {11.77 / 41.77 / 64.16} & {35.91} \\ \hline
\multicolumn{1}{c|}{} & \multicolumn{1}{c|}{32} & {6.95 / 20.31 / 33.21} & {6.00 / 23.81 / 39.27} & \multicolumn{1}{c|}{{21.59}} & {17.04 / 39.60 / 50.88} & {13.63 / 45.27 / 66.11} & {\textbf{38.75}} \\
\multicolumn{1}{c|}{} & \multicolumn{1}{c|}{64} & {7.87 / 24.25 / 35.86} & {7.01 / 23.75 / 40.09} & \multicolumn{1}{c|}{{\textbf{23.14}}} & {15.04 / 34.73 / 50.66} & {13.01 / 45.58 / 66.68} & {37.62} \\
\multicolumn{1}{c|}{} & \multicolumn{1}{c|}{128} & {7.96 / 21.87 / 34.31} & {6.08 / 24.01 / 40.77} & \multicolumn{1}{c|}{{22.50}} & {15.27 / 34.29 / 48.89} & {12.79 / 44.47 / 66.50} & {37.04} \\
\multicolumn{1}{c|}{\multirow{-4}{*}{\begin{tabular}[c]{@{}c@{}}Mini-batch \\ Size\end{tabular}}} & \multicolumn{1}{c|}{256} & {7.69 / 22.60 / 35.13} & {6.18 / 23.93 / 39.58} & \multicolumn{1}{c|}{{22.52}} & {16.59 / 34.07 / 48.45} & {12.43 / 41.90 / 66.42} & {36.64} \\ \hline
\end{tabular}}
\end{table*}

\subsection{Visual Analysis}
\subsubsection{Qualitative Results of Image-text Retrieval}
We selected two representative images and sentences and visualized the retrieval results of top-5, as shown in Fig. \ref{fig:fig6}. Sentence retrieval refers to the use of an image as a query to search for matching text, similar to image retrieval, which uses text as a query to retrieve matching images. Observing Fig. \ref{fig:fig6}(a) and (b), the retrieval results of the top 5 are highly similar and hard to distinguish. Instead, our method can get more accurate retrieval results by enhancing the visual and textual representations. In sentence retrieval, it is challenging to retrieve matching sentences for images without significant objects. In image retrieval, our model has better retrieval performance regardless of retrieving images with or without salient objects. In summary, our model can better achieve bidirectional retrieval of images and text.

\subsubsection{Statistical Analysis of Embedding Distances}
In the high-dimensional embedding space, there is some connection between the distances of different types of embeddings. Using the ROAM module, our proposed DOVE method adaptively adjusts the distances between the final visual and textual embeddings. To express this layer of relationship, we count the Euclidean distances between different types of embeddings corresponding to all positive sample pairs from the RSITMD test set, as shown in Fig. \ref{fig:fig9}. We counted multiscale visual embeddings $\bm{V}_{M}$, regional visual embeddings $\bm{V}_{R}$, word-level textual embeddings $\bm{T}_G$, final visual embeddings $\bm{V}_{M R}$ and textual embeddings $\bm{T}_{R G}$ (combine with Fig. \ref{fig:fig2}), among which regional visual embeddings changes the distance only slightly by linear transformation as orientation. We find that the distance between the final visual embedding and the text embedding is smaller than the distance between the multiscale visual embedding and the word-level textual embedding ($d(\bm{V}_{M R},\bm{T}_{R G}) < d(\bm{V}_{M},\bm{T}_{G})$); the distance from the final visual embedding to the regional visual embedding is smaller than the distance from the multiscale visual embedding to the regional visual embedding ($d(\bm{V}_{R},\bm{V}_{MR}) < d(\bm{V}_{R},\bm{V}_{M})$); the distance from the final textual embedding to the regional visual embedding is smaller than the distance from the word-level text embedding to the regional visual embedding ($d(\bm{V}_{R},\bm{T}_{R G}) < d(\bm{V}_{R},\bm{T}_{G})$); specifically, regional visual features guide the representation of the text by adjusting the spatial distance more slightly. The above results demonstrate that in our DOVE model, the regional visual embedding plays an oriented role in the latent semantic space; the multiscale visual embedding and word-level text embedding serve as an external global visual-semantic constraint that holds the distance between the final visual and textual embeddings.

\begin{figure*}[ht]
  \centering
  \includegraphics[width=\linewidth]{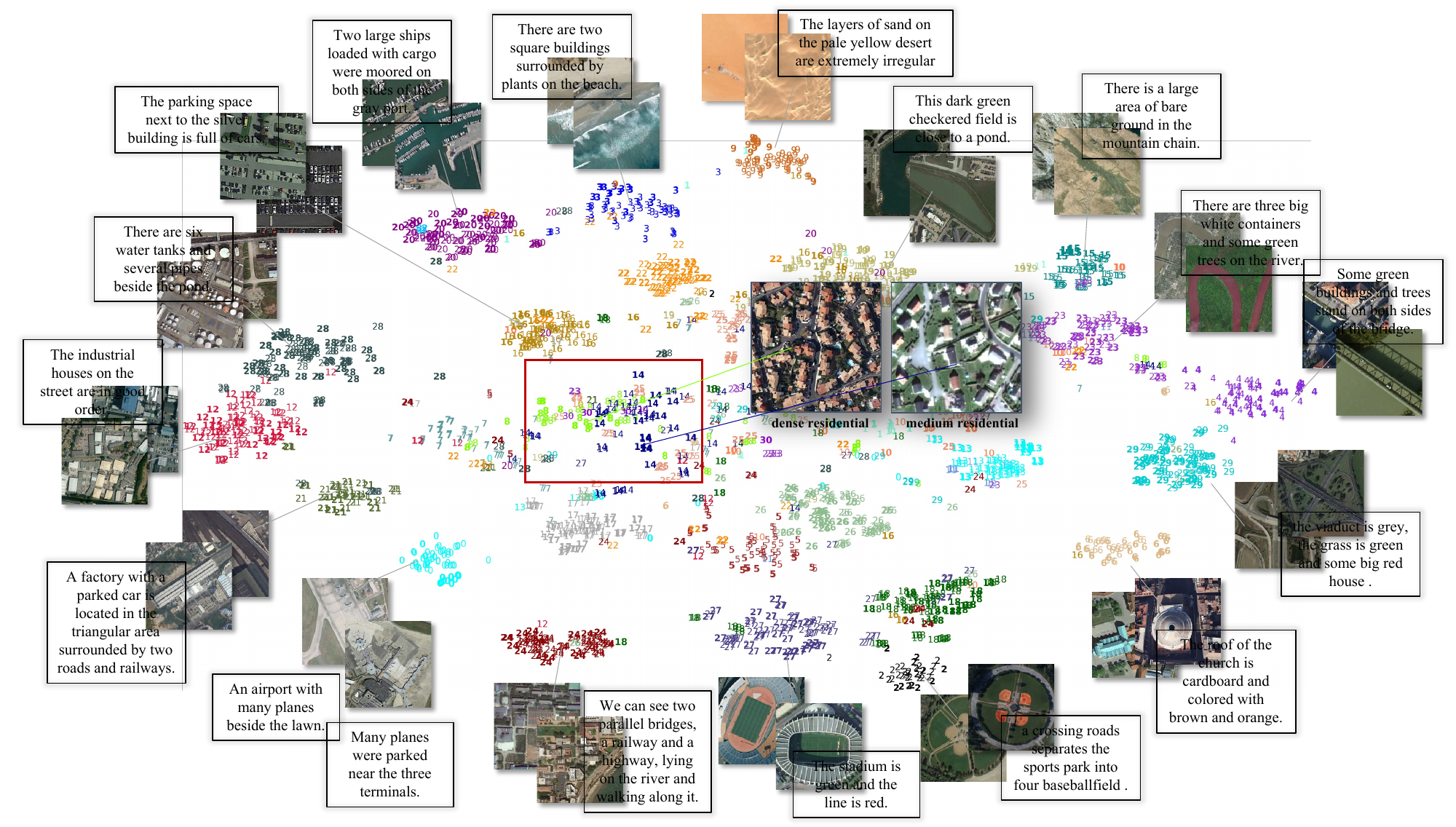}
  \caption{Visualization of latent embedding space. Colored numbers represent images or text of different scenes; normal font: final textual embeddings;  bold font: final visual embeddings.}
  \label{fig:fig7}
\end{figure*}

\subsubsection{Visualization of Latent Embedding Space}
To visually evaluate the contribution of the DOVE model to the final visual and textual representations, we used the t-SNE \cite{van2008visualizing} to visualize the final visual and textual embeddings in latent embedding space, as shown in Fig. \ref{fig:fig7}. It can be observed that the latent embedding space visualization of remote sensing image-text retrieval presents a cluster-like distribution according to different scene types, and semantically similar images or sentences are close in latent embedding space. However, ``\emph{dense residential}'' and ``\emph{medium residential}'' differ only in housing density, and it is difficult to define their scene categories. This case tends to cause the model to learn an incorrect visual representation, and aggravate the visual-semantic imbalance. It can be found that matching images and sentences are close to each other and as far away from the mismatching ones as possible. Many mismatching image-text pairs of different scene categories are close to each other, such as ``\emph{farmland}'', ``\emph{park},'' and ``\emph{resort},'' which is an apparent inter-class similarity that can easily lead to visual-semantic imbalance. Similarly, the second example demonstrates the same conclusion. The above results reflects that our method can identify most of the scenes better, but there are still some scenes that are harder to distinguish.

\subsubsection{Exploring Semantic Localization}
\begin{figure}[t]
  \centering
  \includegraphics[width=\linewidth]{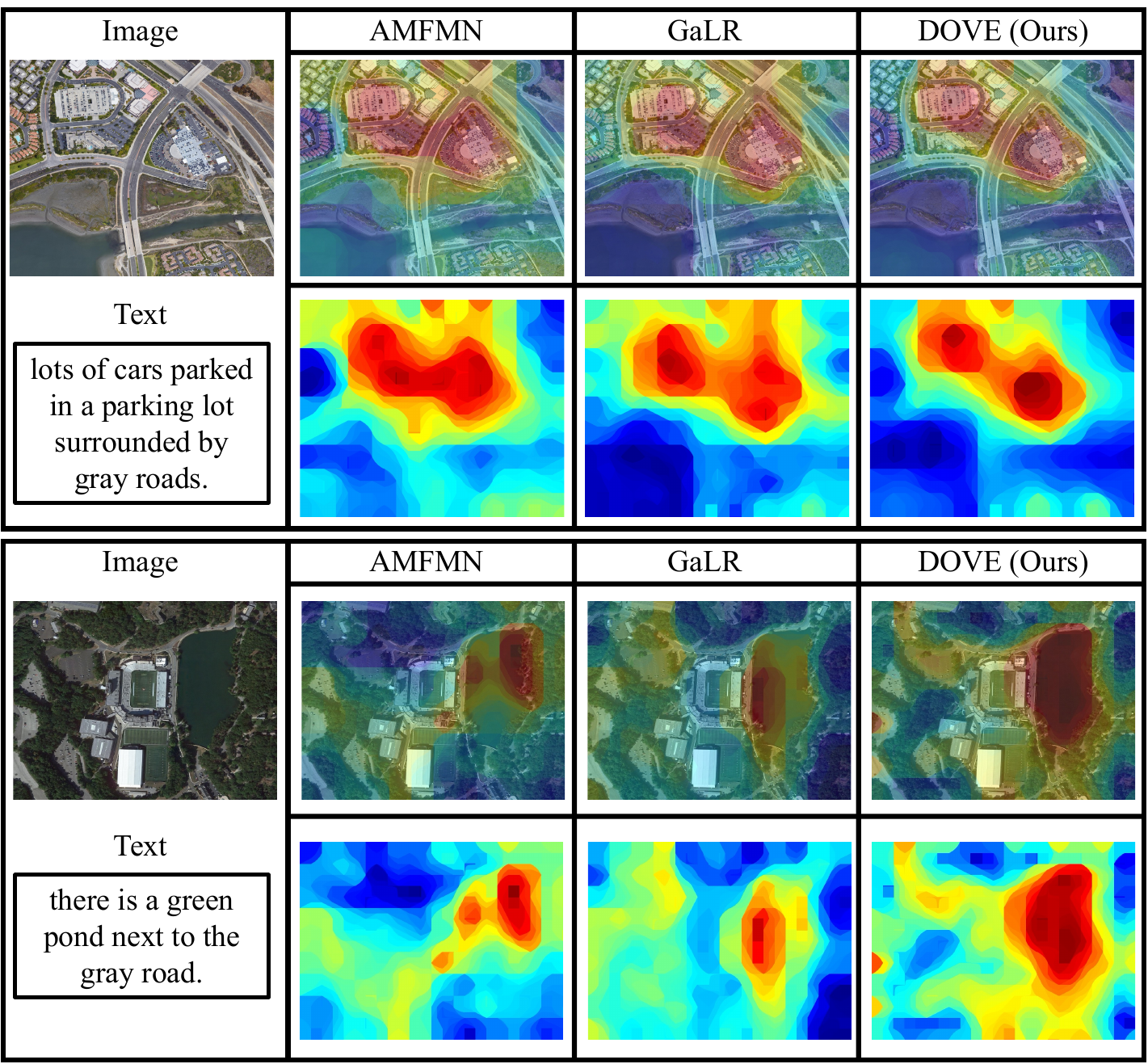}
  \caption{Results of remote sensing image-text retrieval methods on semantic localization. First row: semantic localization result; second row: heat map, where colors closer to red indicate more semantic relevance and blue is the opposite.}
  \label{fig:fig8}
\end{figure}

Semantic localization \cite{yuan2022learning} refers to using text as query to obtain the best matching location in large-scale remote sensing images. We compared our method with AMFMN \cite{yuan2022exploring} and GaLR \cite{yuan2022remote} for semantic localization, with results as shown in Fig. \ref{fig:fig8}. For example, the text ``\emph{lots of cars parked in a parking lot surrounded by gray roads}'' retrieved the most semantically relevant areas in the image. From the semantic localization results, the DOVE model has a more realistic localization effect, whose relevant regions have more precise localization boundaries. Heat map results show that the DOVE model has more accurate edges in the dense region of ``\emph{cars}'' than AMFMN and GaLR, which indicates that our method has more accurate location identification in semantic localization. The blue part in the heat map shows model's filtering ability for redundancy. Compared with the other two methods, the DOVE model can identify the redundant features more accurately. The experiments qualitatively illustrate the superiority of our method in visual semantic understanding.

\section{CONCLUSION}
In this paper, we proposed the DOVE model to solve the visual-semantic imbalance in remote sensing image-text retrieval. The ROAM module adaptively adjust the distance between final visual embedding and final textual embedding to mine the intrinsic connection between vision and language. To enhance textual representation, the DTGA model learns a better textual representation using forward and backward contextual semantics. A global visual-semantic constraint acts as an external constraint for the final visual and textual representations and reduce single visual dependency. Experiments demonstrated the effectiveness of DOVE, which outperformed state-of-the-art methods on the RSICD and RSITMD datasets.

Mining the semantics of remote-sensing images is a valuable and necessary task. In the future, we will continue to explore further applications and enhancements of image-text retrieval in remote sensing. A more integrated and unified model, adapted to the special environment of remote sensing, might be needed for the current remote sensing image-text retrieval.

\bibliographystyle{IEEEtran}

\bibliography{references}

\begin{IEEEbiography}[{\includegraphics[width=0.85in ,clip,keepaspectratio]{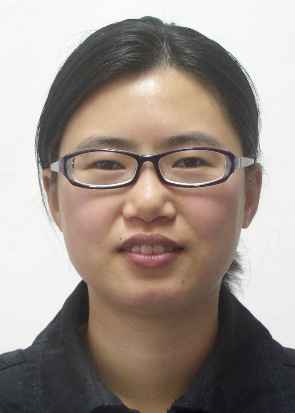}}]{Qing Ma}
received the B.S. and M.S. degrees from Beijing Normal University, Beijing, China, in 2002 and 2005, respectively, and the Ph.D. degree from the Zhejiang University of Technology, Hangzhou, China, in 2021. Since 2005, she has been on the faculty of the College of Science, Zhejiang University of Technology. Her research interests include cross-modal retrieval and computer vision.
\end{IEEEbiography}

\begin{IEEEbiography}[{\includegraphics[width=0.85in ,clip,keepaspectratio]{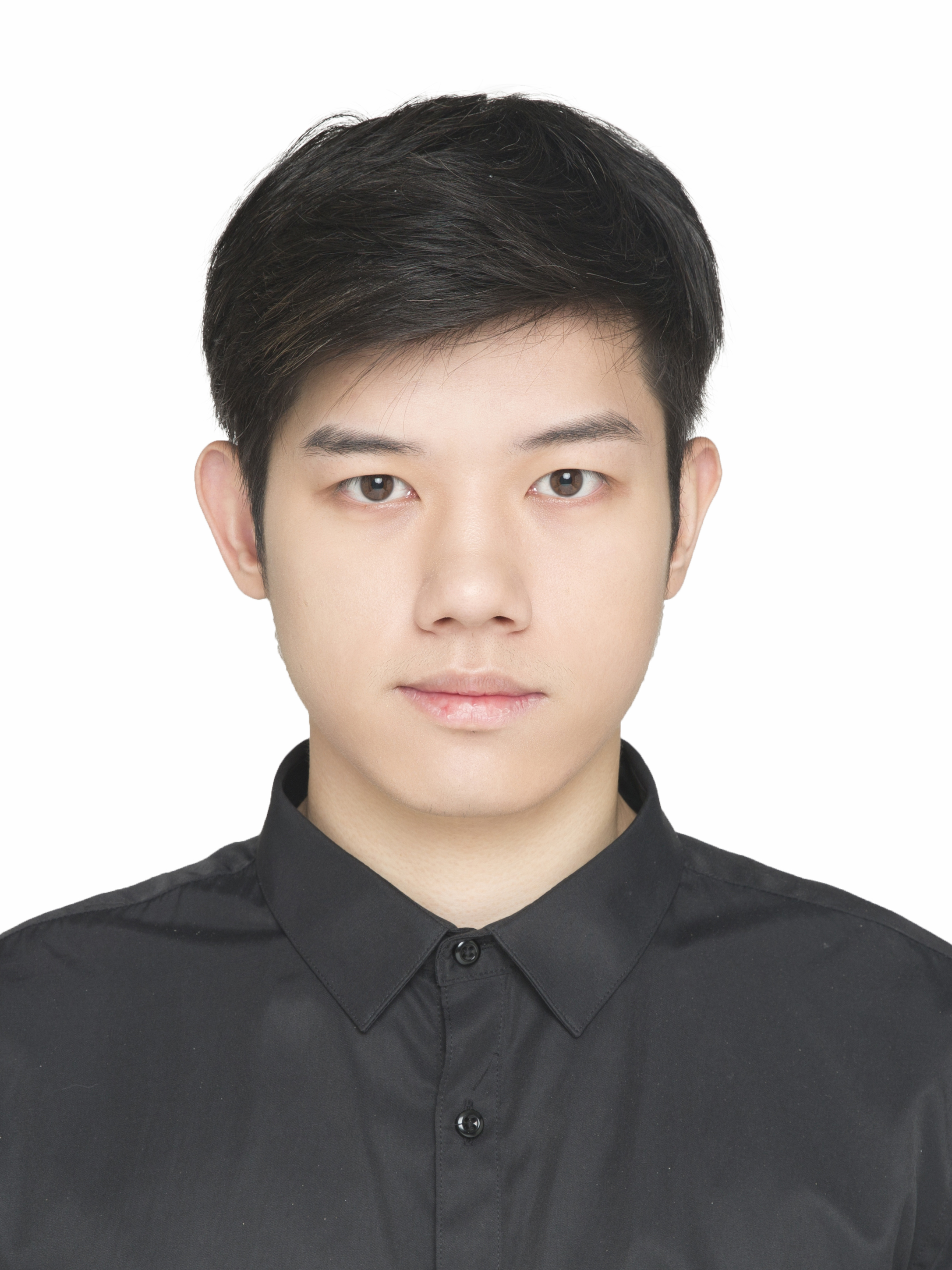}}]{Jiancheng Pan}
(Student Member, IEEE) received the B.E. degree from Jiangxi Normal University, Nanchang, China, in 2022. He is currently pursuing the M.E. degree with the Zhejiang University of Technology, Hangzhou, China. 

His research interests include but are not limited to  Cross-modal Retrieval, Vision-Language Models, and AI for Science.
\end{IEEEbiography}

\begin{IEEEbiography}[{\includegraphics[width=0.85in ,clip,keepaspectratio]{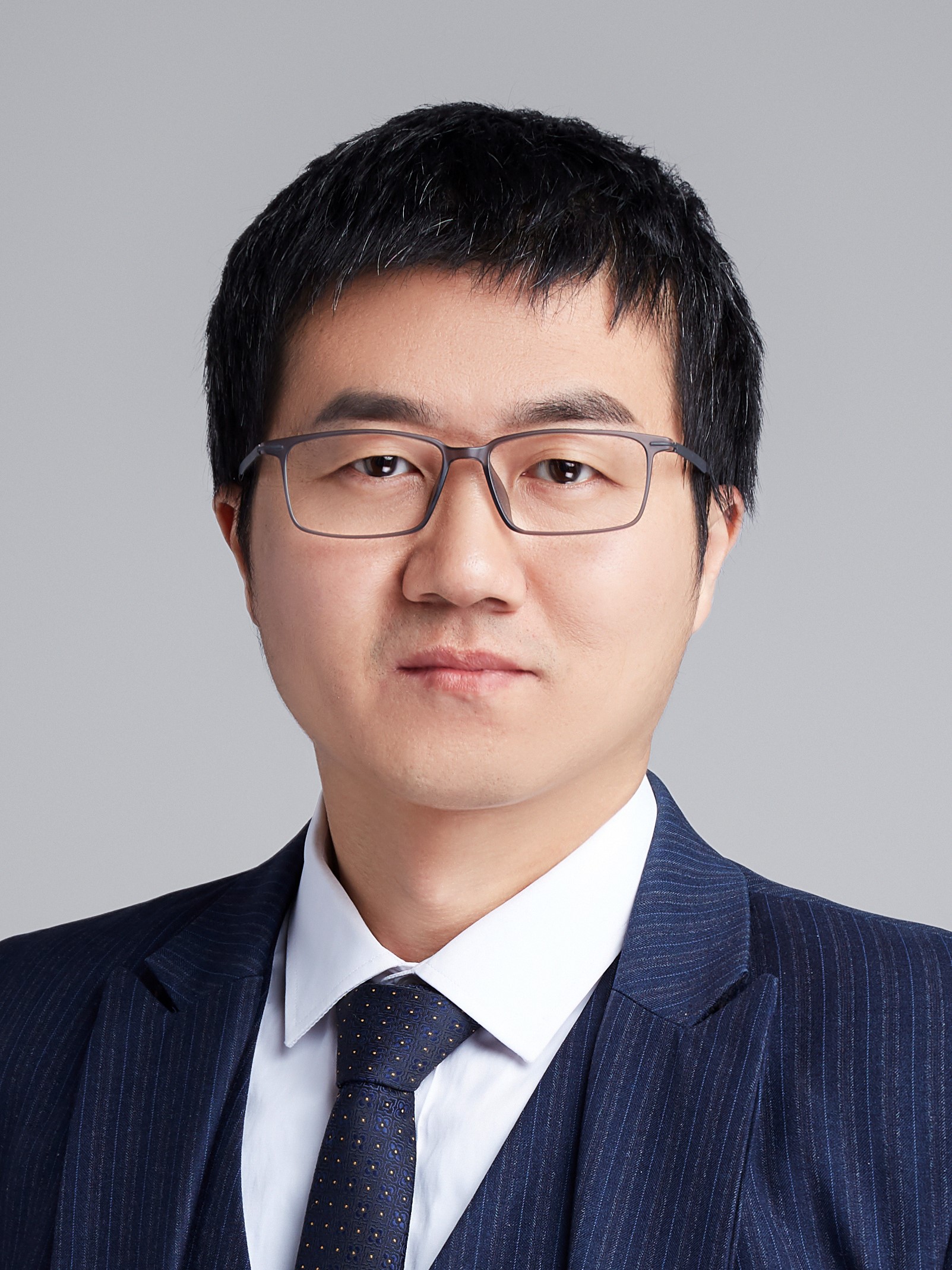}}]{Cong Bai}
(Member, IEEE) received the B.E. degree from Shandong University, Jinan, China, in 2003, the M.E. degree from Shanghai University, Shanghai, China, in 2009, and the Ph.D. degree from the National Institute of Applied Sciences, Rennes, France, in 2013.

He is a Professor with the College of Computer Science and Technology, Zhejiang University of Technology, Hangzhou, China. His research interests include computer vision and multimedia processing.
\end{IEEEbiography}

\end{document}